\documentclass{article}

\PassOptionsToPackage{numbers, compress}{natbib}

    \usepackage[preprint]{neurips_2025}

\usepackage[utf8]{inputenc} 
\usepackage[T1]{fontenc}    
\usepackage{hyperref}       
\usepackage{url}            
\usepackage{booktabs}       
\usepackage{amsfonts}       
\usepackage{nicefrac}       
\usepackage{microtype}      
\usepackage{xcolor}         
\usepackage{amsmath}
\usepackage{amsthm}
\usepackage{amssymb}
\usepackage{changepage}
\usepackage{bbding}
\usepackage{gensymb}

\usepackage{graphicx}
\usepackage{subcaption}
\usepackage{caption}
\usepackage{bm}
\usepackage{titlesec}

\usepackage{todonotes}
\usepackage{enumitem}
\usepackage{cleveref}

\title{TanDiT: Tangent-Plane Diffusion Transformer for High-Quality 360$^{\bm{\circ}}$ Panorama Generation}

\author{%
  Hakan Çapuk$^*$ \\
  Koç University \\
  \And
  Andrew Bond$^*$ \\
  Koç University \\
  \And
  Muhammed Burak Kızıl \\
  Koç University \\
  \And
  Emir Göçen \\
  Koç University \\
  \And
  Erkut Erdem \\
  Hacettepe University \\
  \And
  Aykut Erdem $^\dagger$ \\
  Koç University \\
}

\begin{document}

\def\thefootnote{*}\footnotetext{Equal Contribution}\def\thefootnote{\arabic{footnote}}
\def\thefootnote{$\dagger$}\footnotetext{Corresponding Author}\def\thefootnote{\arabic{footnote}}

\maketitle

\begin{figure}[ht]
    \label{fig:teaser}
    \centering
    \includegraphics[width=\textwidth]{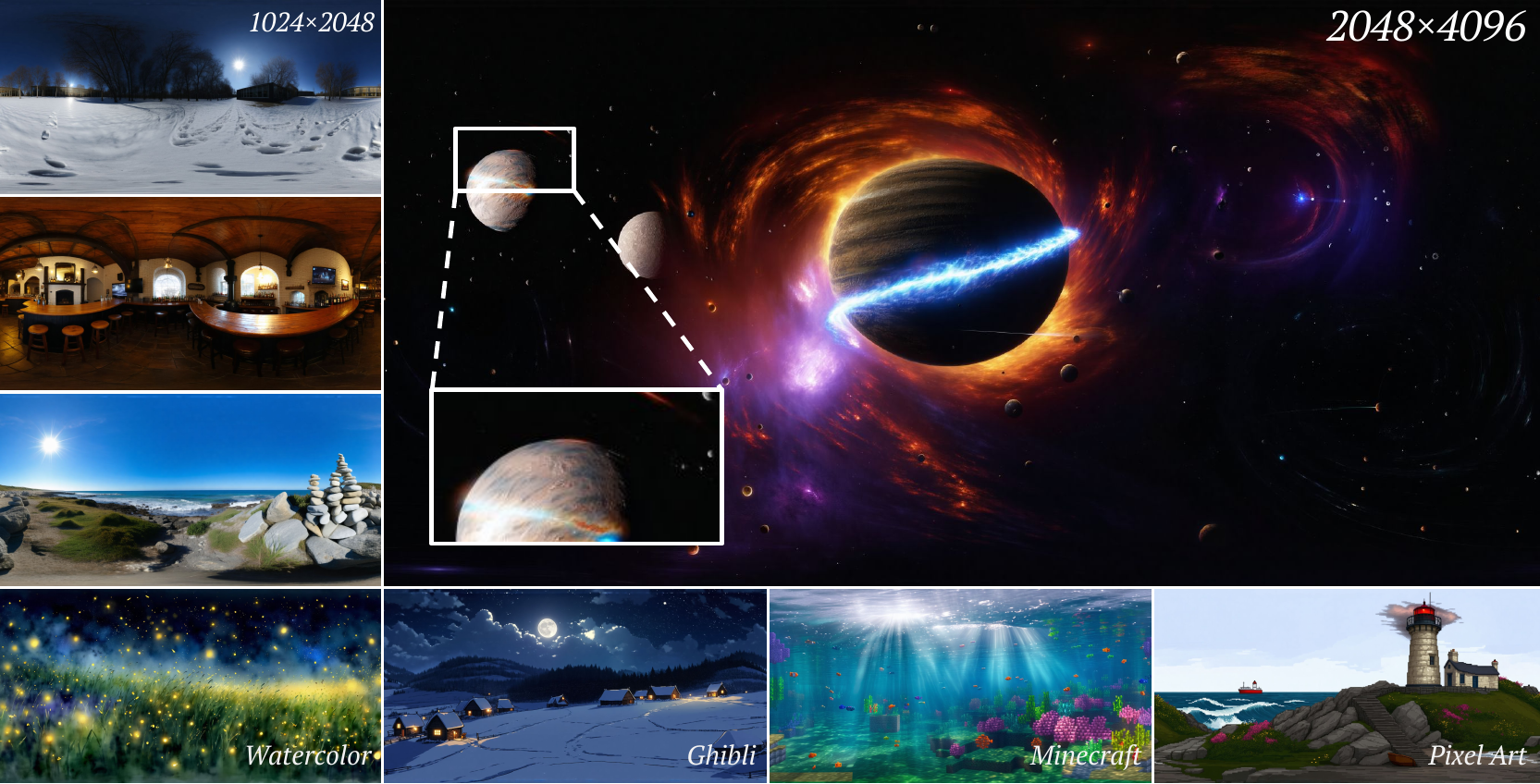}
    \caption{\textbf{Panoramic samples generated by our TanDiT model at varying resolutions and styles.} TanDiT produces visually compelling and seamlessly consistent 360$^\circ$ panoramas across various resolutions, even at 4K, and diverse visual styles. By decomposing spherical images into multiple tangent-plane (perspective) views, our model seamlessly integrates powerful off-the-shelf diffusion transformers and super-resolution methods, enabling flexible and high-quality panorama generation.}
    \label{fig:teaser}
\end{figure}

\begin{abstract}
Recent advances in image generation have led to remarkable improvements in synthesizing perspective images. However, these models still struggle with panoramic image generation due to unique challenges, including varying levels of geometric distortion and the requirement for seamless loop-consistency. To address these issues while leveraging the strengths of the existing models, we introduce TanDiT, a  method that synthesizes panoramic scenes by generating grids of tangent-plane images covering the entire 360$^\circ$ view. Unlike previous methods relying on multiple diffusion branches, TanDiT utilizes a unified diffusion model trained to produce these tangent-plane images simultaneously within a single denoising iteration. Furthermore, we propose a model-agnostic post-processing step specifically designed to enhance global coherence across the generated panoramas. To accurately assess panoramic image quality, we also present two specialized metrics, TangentIS and TangentFID, and provide a comprehensive benchmark comprising captioned panoramic datasets and standardized evaluation scripts. Extensive experiments demonstrate that our method generalizes effectively beyond its training data, robustly interprets detailed and complex text prompts, and seamlessly integrates with various generative models to yield high-quality, diverse panoramic images.
\end{abstract}

\section{Introduction}

Text-to-image diffusion models have recently demonstrated remarkable capabilities in generating high-quality, diverse perspective images. However, extending these successes to the domain of 360$^\circ$ panoramic generation remains a significant challenge. Panoramic scenes present unique difficulties: they have significant distortions in polar regions and require global spatial coherence across a continuous field of view, including between the left and right edges of the image. Existing approaches to panorama generation attempt to overcome these challenges using multi-branch denoising strategies, stitching heuristics, or blending schemes. While these methods offer partial solutions, they often struggle with spatial discontinuities, artifacts in the polar regions, or lack the flexibility to generalize across resolutions and styles. Some training-free approaches such as SphereDiff and CubeDiff use tangent or cubemap projections, but lack integrated generation pipelines that ensure local continuity and global coherence in a unified framework.

In this work, we introduce TanDiT (Tangent-Plane Diffusion Transformer), a new framework for high-quality panoramic image synthesis. TanDiT decomposes a 360$^\circ$ scene into a structured grid of tangent-plane perspective views. Unlike prior work that generates each view independently, TanDiT employs a single, transformer-based diffusion model to generate the entire grid in one pass, preserving spatial adjacency and enabling information sharing between overlapping regions. This leverages the strengths of modern Diffusion Transformers (DiTs), whose attention mechanisms are well-suited to modeling complex spatial relationships across structured inputs. Additionally, we propose an equirectangular-conditioned refinement step, in order to apply additional processing like super-resolution while still maintaining global consistency, and ensure that the generated tangent planes better blend together seamlessly, while keeping the core structure of the scene the same.

We also introduce two new evaluation metrics tailored for panoramic image quality: TangentIS and TangentFID, which compute performance over extracted tangent views to better capture fidelity and diversity across the entire spherical field of view. These are part of our new proposed evaluation suite for panoramic image generation, including a collection of text captions for multiple commonly used panoramic image datasets which have no publicly available captions. We also develop a robust evaluation suite encompassing all commonly used metrics for this task and including our two new proposed metrics. Extensive experiments demonstrate that TanDiT achieves state-of-the-art results across multiple quantitative benchmarks and qualitative assessments, while remaining robust to diverse styles and challenging textual prompts.

In summary, our contributions are: (i) A novel panoramic image generation framework leveraging the capabilities of Diffusion Transformers, and the perspective nature of tangent planes, in order to generate high-quality and diverse panoramic images. (ii) An equirectangular-conditioned refinement strategy to ensure boundary consistency and image realism, handle any inconsistencies between the generated tangent planes, and allow for applications like super-resolution much more smoothly than other approaches. (iii) Two new panoramic-specific evaluation metrics and a benchmarking suite that enable more accurate and fair comparisons between panoramic generation models.

\section{Related Work}
\label{related_works}


\textbf{Image Generation.} Recently, Diffusion Models have come to dominate the domain of conditional image generation. Stable Diffusion~\cite{rombach2022high} and DALL-E 2~\cite{dalle2} leverage a U-Net based architecture to predict noise during a denoising process in the latent space, mapping random noise to latent representations to be decoded into images. More recently, Stable Diffusion 3~\cite{sd3} and Flux~\cite{flux} replaced the traditional U-Net with a transformer-based backbone, offering improved scalability and the ability to capture global context more effectively.

\textbf{Panorama Generation.} MultiDiffusion~\cite{multidiffusion} divides the panoramic image into overlapping patches that are denoised independently and blended to reduce seams. StitchDiffusion~\cite{stitchdiffusion} extends MultiDiffusion by introducing additional overlapping regions and explicitly stitches them to ensure left-right consistency. PanFusion~\cite{panfusion} proposes a dual branch strategy in which one branch generates multiview perspective images and the other serves as a global canvas for the panorama in the form of an equirectangular projection (ERP) image. Diffusion360~\cite{diffusion360} and PanoDiff~\cite{panodiff} employ a circular blending strategy to maintain consistency at opposite opposite panorama boundaries. MVDiffusion~\cite{mvdiffusion} takes eight perspective views and projects them into a closed-loop panoramic image. CurvedDiffusion~\cite{curveddiffusion} introduces lens geometry-based conditioning to enable full 360° panorama generation. PanoFree~\cite{panofree}, like MultiDiffusion, follows a training-free paradigm, iteratively generating multiview images through warping and inpainting. CubeDiff~\cite{cubediff} uses an outpainting strategy by starting from a single cubemap face and predicting the remaining faces to reconstruct the full panorama. 360DVD \cite{360dvd} introduces an adapter to extend pre-trained T2V models for panoramic video generation. 4K4DGen \cite{4k4dgen} utilizes dynamic Gaussian Splatting to transform a panoramic image into a 4D scene. Finally, SphereDiff~\cite{spherediff} modifies the MultiDiffusion approach along with the use of tangent images to get a training-free method for image and video generation. A summary of the key differences among prior 360° image generation methods is provided in Table~\ref{tab:method_comparison}. Unlike previous works, TanDiT is the only method that combines tangent-plane representations with consistency-aware generation and arbitrary resolution support in a single diffusion loop, while also generalizing effectively to out-of-domain prompts.

\begin{table}[!t]
    \caption{\textbf{Comparison of recent panorama generation methods.} Models are compared based on image type, open-source availability, consistency, training requirements, resolution flexibility, and ability to generate high-quality out-of-domain images. TanDiT is the only method that uses tangent-plane representations with consistency and arbitrary resolution in a single diffusion loop.}
    \centering
    \resizebox{\textwidth}{!}{
    \begin{tabular}{@{}l@{$\;\;\,$}l@{$\;\;\,$}c@{$\;\;\,$}c@{$\;\;\,$}c@{$\;\;\,$}l@{$\;\;\,$}c@{}}
        \midrule
         \textbf{Model} & \textbf{Image Type} & \textbf{$\#$ Generations} & \textbf{Consistent} & \textbf{Training-Free} & \textbf{Resolution} & \textbf{OOD} \\
         \midrule
         PanFusion~\cite{panfusion} & ERP & 1 & \Checkmark & \XSolidBrush & Fixed & \XSolidBrush \\
         MultiDiffusion~\cite{mvdiffusion}  & Wide Perspective & 1 & \Checkmark & \Checkmark & Arbitrary & \Checkmark \\
         StitchDiffusion~\cite{stitchdiffusion}  & ERP & 1 & \Checkmark & \XSolidBrush & Arbitrary & \XSolidBrush \\
         Diffusion360~\cite{diffusion360}  & ERP & 1 & \Checkmark & \XSolidBrush & Arbitrary & \XSolidBrush \\
         CubeDiff~\cite{cubediff} & Cubemap & 5 & \XSolidBrush & \XSolidBrush & Fixed & \Checkmark \\
         SphereDiff~\cite{spherediff}  & Tangent Plane & 89 & \XSolidBrush & \Checkmark & Arbitrary & \Checkmark \\
         \midrule
         TanDiT (Ours) & Tangent Plane & 1/2 & \Checkmark & \XSolidBrush & Arbitrary & \Checkmark \\
         \midrule
    \end{tabular}}
    \label{tab:method_comparison}
\end{table}



\section{Methodology}
\label{methodology}
\subsection{Preliminary}
\textbf{Diffusion Models.} Diffusion models \cite{ho2020denoising, sohl2015deep} are a class of generative models which synthesize data by starting from random noise and denoising it through a learned stochastic process. During training, the model learns a prior over noise conditioned on the timestep by optimizing a noise prediction objective: noise is added to an image, and the model is trained to predict the noise. Many modern diffusion models follow the latent diffusion paradigm \cite{rombach2022high}, which applies the diffusion process in a lower-dimensional latent space rather than pixel space, significantly improving efficiency. This latent space is obtained using a variational autoencoder (VAE) \cite{vae}, with compression ratios in the range of 8$\times$ to 16$\times$. More recently, an alternative training paradigm known as flow matching \cite{flow_matching, rectified_flows} has gained attention. Flow matching formulates the training process as an ordinary differential equation, learning a vector field that transforms the noise distribution into the data distribution. 
\begin{equation}
    \mathcal{L}_{\text{CFM}} = \mathbb{E}_{t, p_t(\mathbf{z}|\boldsymbol{\epsilon}), p(\boldsymbol{\epsilon})} || v_{\theta}(\mathbf{z}, t) - u_t(\mathbf{z}, \boldsymbol {\epsilon})||_{2}^{2}, \quad \mathbf{z}_{t} = (1 - t) \mathbf{x}_{0} + t \boldsymbol{\epsilon}
\end{equation}
where $\epsilon$ represents standard gaussian noise, and $\mathbf{z}_t$ is the latent at time $t$. This underlies recent diffusion models such as Stable Diffusion 3 \citep{sd3} and Flux \citep{flux}.

\textbf{Diffusion Transformers.} Early diffusion models, such as DDPM \citep{ho2020denoising} and Stable Diffusion \citep{rombach2022high}, typically relied on U-Net-based architectures \citep{unet}. More recently, Diffusion Transformers (DiTs)~\citep{peebles2022dit} have emerged as the dominant architecture for diffusion models. While architectural details may vary, most DiT models adopt a double-stream design, where image and text features are processed separately within each block and then integrated through a cross-attention mechanism before continuing independently to the next layer. Some models also include single-stream blocks, where all tokens are merged and processed jointly. Stable Diffusion 3 \citep{sd3} employs only double-stream blocks, whereas Flux \citep{flux} combines double-stream layers with final single-stream layers for joint reasoning.
DiT models have been shown to have very strong correlations in spatially close pixels \citep{makeanything}, a convenient property which we use for our approach.

\textbf{Gnomonic Projection and Tangent Planes.}
In our approach, rather than generating the entire panoramic image-which contains varying amount of distortion arising from the characteristics of equirectangularly projected images, we generate a set of tangent planes that can later be projected onto the panorama. This allows us to better leverage the generative capabilities of diffusion models, which perform more effectively on undistorted, perspective-aligned views. Tangent planes are obtained via the gnomonic projection, which maps a point $(\theta, \phi)$ on the surface of a sphere to a point $(x,y)$ on a plane tangent to the sphere at a reference point $(\theta_t, \phi_t)$. 
Further details on the projection process and the properties of tangent planes are provided in Appendix~\ref{sec:tangent-plane}.
\subsection{Model Design}
\label{sec:model_design}
\begin{figure}[!t]
    \centering
    \includegraphics[width=\textwidth]{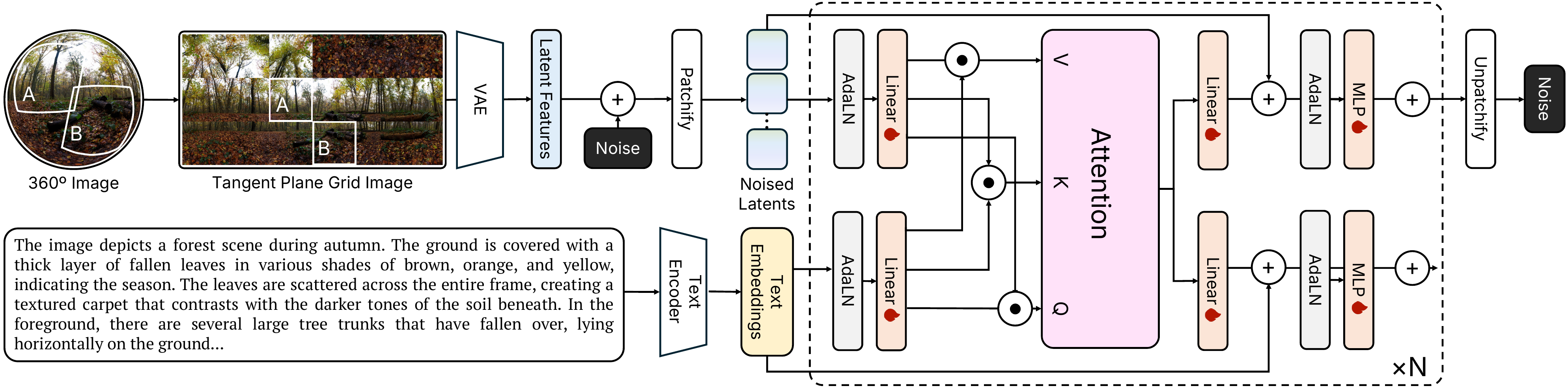}
    \caption{\textbf{Overview of the TanDiT training pipeline.} Our method starts by decomposing a 360$^\circ$ panoramic image into a structured grid of tangent-plane projections via gnomonic projection. These projections are arranged into a single coherent grid image, ensuring adjacent placement of overlapping regions for spatial consistency. Given a dense textual caption describing the scene, the model is trained to reconstruct this grid using a standard denoising diffusion objective in the latent space. }
    \label{fig:model_diagram}
\end{figure}

A key challenge in generating multiple tangent planes of a panoramic scene simultaneously is ensuring mutual consistency and preserving the correct spatial relationships between them. The relationship between two tangent planes is determined entirely by their positions in the equirectangular projection. For instance, two neighboring tangent planes may have significant overlap, while distant ones may share no common content. designing a generation process that respects the geometry and spatial continuity of the underlying 360$^\circ$ scene is nontrivial.  To address this, we draw inspiration from the ``serpentine'' image arrangement proposed in \cite{makeanything}, originally used to organize temporally ordered frames into a 2D grid. Specifically, we construct a grid of tangent views such that neighbors in the equirectangular projection remain adjacent in the grid layout. Rather than generating tangent planes step-by-step, we train the model to generate the entire grid jointly. This design enables the DiT to exploit local context across views, promoting globally coherent generation. Figure~\ref{fig:model_diagram} summarizes the full DiT training loop on a grid of tangent planes.

Once the tangent planes are extracted and arranged into a grid, we train the model using the standard DiT training objective. Since the VAE-based latent space preserves spatial structure, the DiT can effectively capture and propagate local dependencies across tangent views. In order for these spatial relationships to be informative, it is important to arrange the grids in such a way that spatially close grid images are also spatially close in the 360° image. We give a detailed explanation of our grid construction strategy and specific training setup and hyperparameters in Appendix~\ref{sec:supp_dataset}.



\vspace{-10pt}

\subsection{Equirectangular-Conditioned Refinement}
\label{sec:refinement}
When the tangent-plane grid is reprojected into a single panoramic view via equirectangular projection, inconsistencies can appear in the overlapping regions between neighboring views. These artifacts arise because the diffusion objective on the spatially-separated grid tokens does not explicitly enforce overlap coherence, leading to visible seams and unnatural transitions. To correct this, we add a model-agnostic refinement step that harmonizes overlaps and boosts global realism.

Specifically, we first reconstruct an ERP image by “stitching” the generated tangent planes into a full 360° panorama, then encode it with the VAE’s encoder to obtain
\begin{equation}
    z_{\text{pano}} = \text{Encoder}\big(\text{ERP}(\{\text{tangent views}\})\big),
\end{equation}
which differs from the latent of the original grid image. We then perturb this panoramic latent with noise at a high diffusion timestep $T_{\text{high}}\sim800$ and denoise it using the pretrained DiT, conditioned on the original text prompt $c$ and the timestep $t$:
\begin{equation}
    \mathcal{D}_{\theta} (\mathbf{z}_{\text{pano}} + \sigma_{T_{\text{high}}} \cdot \epsilon, c, t) \xrightarrow[t \rightarrow 0]{} 
    \mathbf{z}_{\text{refined}}
\end{equation}
Finally, the refined latent is decoded back to pixel space:
\begin{equation}
    I_{\text{final}} = \text{Decoder}(\mathbf{z}_{\text{refined}})
\end{equation}
Here, $\mathcal{D}_{\theta}$ denotes the pre-trained DiT model, $\sigma_{T_{\text{high}}}$ is the noise scale at timestep $T_{\text{high}}$, and $\epsilon \sim \mathcal{N}(0, I)$. By using a high timestep, we wash out high-frequency mismatches while preserving the scene's overall layout, letting the DiT re-sculpt smooth, prompt-aligned panoramas from the low-frequency structure encoded in $\mathbf{z}_{\text{pano}}$.

\begin{figure}
    \centering
    \includegraphics[width=\textwidth]{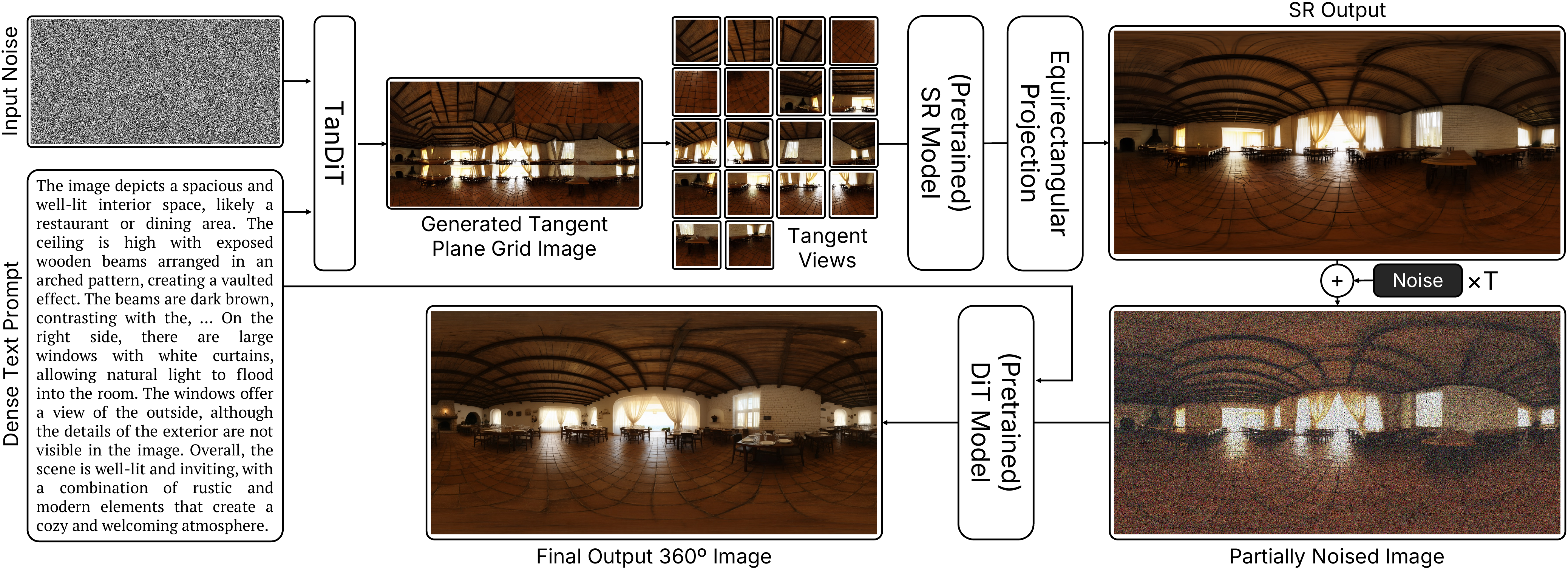}
    \caption{
 \textbf{Overview of the TanDiT inference pipeline.} At inference time, TanDiT first generates a grid of tangent views conditioned on a text prompt. These tangent views are enhanced using a super-resolution module and then reprojected to form an intermediate equirectangular panorama. To further improve global coherence and visual quality, the latent representation of this panorama is perturbed with noise and refined by a pre-trained DiT model, conditioned on the same text input, producing the final high-resolution 360$^\circ$ image.}
    \label{fig:inference_pipeline}
\end{figure}

However, this refinement step introduces loop-inconsistency within the panorama, as there is insufficient information flow between the right and left edges of the image during denoising. In order to mitigate this, we adapt a circular padding strategy \cite{diffusion360, panodiff} to ensure that the edge regions receive meaningful context at each denoising step, thereby enabling the generated panorama to be loop-consistent. Formally, the circular padding is defined as
\begin{equation}
    \hat{\mathbf{z}}_{i,j} = \mathbf{z}_{i, j - p\text{ (mod $W$)}}
\end{equation}
where $\hat{\mathbf{z}}_{i,j}$ is the $(i,j)$-th position of the padded latent, $p$ is the amount of padding on one side, and $W$ is the width of the latent.
Finally, when super-resolution is applied to the intermediate equirectangular image, its resolution increases to either 1024$\times$2048 or 2048$\times$4096 pixels. Most DiTs, including Stable Diffusion 3 \cite{sd3}, struggle to denoise such high-resolution rectangular inputs, often leading to artifacts and textural degradation near the image boundaries. To address this, we divide the image into 1024$\times$1024 pixels square patches and denoise each patch seperately. This strategy enables better utilization of the model’s capacity while allowing generation of high-quality images with a 1:2 aspect ratio. An overview of this inference pipeline is shown in Figure~\ref{fig:inference_pipeline}.

\section{Experimental Setup}
\subsection{Dataset}
\label{sec:dataset}
We use a combination of 3 different datasets: Matterport3D \cite{matterport3d} ($\sim 10$K images) , Polyhaven \cite{polyhaven} ($\sim 750$ images), and Flickr360 \cite{flickr360} ($\sim 3$K images). None of these datasets come with text captions, and so we compute these ourselves. We employ LLava-OneVision \cite{llavaonevision} to produce rich, dense captions for the images in our dataset. We furthermore use a Llama-2 model \cite{llama2} to summarize these images into concise captions, for models which do not have sufficient context length in their text encoder to handle the dense captions, as well as for CLIP Score evaluation \cite{clip}. 
Following \cite{omnifusion}, we create sets of 18 tangent planes for our panoramas, aiming to be able to perform a near-lossless projection back to the panorama. We resize each tangent plane to a resolution of 192$\times$192 pixels, and put the tangent planes in a grid with a layout of 3$\times$6, resulting in 576$\times$1152 pixels grid images. 


\subsection{Evaluation Framework}
While there are some recent works that have explored the setting of panoramic or 360° text-to-image generation, it is still a vastly underexplored domain. We thus develop and propose a new evaluation and benchmark framework, with a new dataset, as well as new metrics that better capture the complexities of a panoramic scene, and a benchmarking suite that provides unified implementations of all relevant metrics for 360° text-to-image generation, including both of our proposed metrics. 

Although several text-to-image models have been proposed for panoramic generation, there is currently no standardized dataset. Existing works typically rely on their own captioning pipelines, which introduces inconsistencies and makes fair comparison difficult, as the quality and style of captions can significantly influence model performance. To address this, we will publicly release our caption set, including both dense and summarized versions, with the goal of establishing a unified benchmark for panoramic text-to-image generation.

\textbf{TangentIS.}
Inception Score (IS) \cite{inception_score} is a commonly used no-reference metric for image quality and diversity. However, it performs very poorly on panoramic images due to the wide aspect ratio (usually 2:1), and the warping in the polar regions. To address this, we develop a modification, which we call TangentIS, based on the idea of tangent planes.
The key advantage of tangent planes is the ability to extract perspective images from a 360° image. This suggests that applying IS to the tangent planes is possible. However, using the same IS score for all of the extracted tangent planes would result in 360° images which attain high inconsistencies between polar regions and the rest of the image having higher IS due to the increase in image diversity. 
To resolve this, we instead compute the IS for each of the 18 tangent planes separately. Simply taking an average of these 18 IS values is also problematic. A model that attains very high IS on the equatorial regions but very low IS on the polar regions would still have a high average, due to the fact that there are twice as many equatorial regions as polar regions. Therefore, we propose to use the lower 95\% confidence bound over these 18 IS values. Specifically, we have
\begin{equation}
    \text{IS}_{\text{tangent}} = \mu_{\text{tangent}} - 1.96 * \frac{\sigma_{\text{tangent}}}{\sqrt{18}}
\end{equation}
where $\mu_{\text{tangent}}, \sigma_{\text{tangent}}$ are the mean and standard deviation of the 18 IS values, respectively. This ensures that even if the equatorial regions are very good, bad polar regions will negatively affect the score quite significantly. Thus, a model must be able to generate good results in all regions of the 360° image in order to achieve the best TangentIS. Further discussion about our choices can be found in Appendix~\ref{sec:metrics_discussion}.

\textbf{TangentFID.}
\label{sec:tangentfid}
For the FID metric \cite{FID}, there does exist a metric which attempts to modify this metric for 360° images, called OmniFID \cite{omnifid}. This metric works by decomposing an equirectangular image into its cubemap representation (bottom, top, and 4 sides), and then computing 3 sets of FID (bottom, top, and middle), using the averaged feature vectors for the middle. These 3 sets of FID are then averaged to produce the final OmniFID score. However, this introduces its own issues. The conversion between equirectangular and cubemap images can introduce distortions near the edge of each of the images, especially for the polar regions. Furthermore, averaging the features in the equatorial region can destroy some context about the individual regions of the image.
Therefore, we propose our own FID metric for 360° images, which we call TangentFID. Similar to our proposed TangentIS, we first start by calculating the FID for each of the 18 tangent planes separately. This time, we apply the upper $95\%$ confidence bound over these 18 values. We also include additional discussion in Appendix~\ref{sec:metrics_discussion} for this metric.

\textbf{Benchmarking Suite.} 
Finally, we release a benchmarking tool which is able to easily and quickly calculate all relevant metrics for 360° images, as well as both of our newly introduced metrics. With a unified implementation of all metrics, our hope is that this will allow for more consistent testing of these models and more reliable results across studies.
\vspace{-10pt}
\section{Results}
\subsection{Quantitative Results}
We evaluate all models on both perspective and panoramic-based metrics. We use the common image generation metrics FID, KID \cite{FID} and IS \cite{inception_score}. CLIP Score \cite{clip} measures the alignment between the generated image and the input text prompt. FAED \cite{faed} uses an autoencoder specially trained on panoramic images, to be able to better extract panoramic-specific features. OmniFID \cite{omnifid} modifies FID to extract the cubemap representation from an ERP image, and takes the average FID over them. Discontinuity Score (DS) \cite{omnifid} measures the existence of inconsistencies and seams in a generated ERP image by using a Scharr kernel. Finally, we discuss TangentIS and TangentFID above in Sec.~\ref{sec:tangentfid}.

Table \ref{tab:main_metrics} contains all results on these metrics. Our model achieves the best or second-best accuracy in most metrics. However, since the IS metric is meant for perspective images, models which produce accurate ERP images attain lower metrics than those that produce only wide perspective images (specifically MultiDiffusion). With CLIP Score, the CLIP encoder also expects perspective images as input, and is unable to process the dense captions used in our generations. As with the IS metric, CLIP Score also assigns higher results to MultiDiffusion since it produces a perspective image, and also uses the exact same captions for generation that the CLIP model sees. Finally, our OmniFID score is very close to both Panfusion and Diffusion360, despite our results being more consistent and detailed, with fewer artifacts in the polar regions, supporting our concerns about the OmniFID metric. 
 \begin{table}[!t]
      \caption{\textbf{Quantitative comparison of panoramic image generation models across standard and proposed metrics.} TanDiT demonstrates consistently strong performance across all metrics, reflecting its ability to generate high-quality, coherent, and well-aligned panoramic images. The best result is highlighted in \textbf{bold}, while the second-best result is indicated with an \underline{underline}.}
     \centering
     \renewcommand{\arraystretch}{1.25}
     \resizebox{\textwidth}{!}{
     \begin{tabular}{l@{$\quad$}c@{$\quad$}c@{$\quad$}c@{$\quad$}c@{$\quad$}c@{$\quad$}c@{$\quad$}c@{$\quad$}c@{$\quad$}c}
          \toprule
          \textbf{Model} & \textbf{FID}$\downarrow$ & \textbf{KID}$\downarrow$ & \textbf{IS}$\uparrow$ & \textbf{CS}$\uparrow$ & \textbf{FAED}$\downarrow$ & \textbf{OmniFID}$\downarrow$ & \textbf{DS}$\downarrow$ & \textbf{TangentFID}$\downarrow$ & \textbf{TangentIS}$\uparrow$ \\
          \midrule
          TanDiT (Ours) & \textbf{32.03} & \underline{0.013} & 4.49 & \underline{23.48} & \textbf{2.56} & 49.62 & \textbf{0.0004} & \textbf{35.39} & \textbf{6.06} \\
          Panfusion & \underline{35.25} & \textbf{0.012} & 4.65 & 23.05 & 2.93 & \underline{48.36} & \underline{0.0007} & 39.57 & \underline{5.68} \\
          StitchDiffusion & 69.68 & 0.040 & \underline{4.92} & 22.09 & 8.46 & 108.46 & 0.0021 & 59.24 & 2.74\\
          Diffusion360 & 42.56 & 0.024 & 3.59 & 21.25 & \underline{2.80} & \textbf{48.32} & \underline{0.0007} & \underline{37.48} & 1.99\\
          MultiDiffusion & 79.35 & 0.050 & \textbf{7.27} & \textbf{26.92} & 4.00 & 88.68 & 0.0050 & 58.57 & 2.37\\
          \bottomrule
          
     \end{tabular}}
     \label{tab:main_metrics}
 \end{table}
 \vspace{-10pt}

 \subsection{Qualitative Results}
\label{sec:qualitative}
\begin{figure}[!t]
    \centering
    \includegraphics[width=\textwidth]{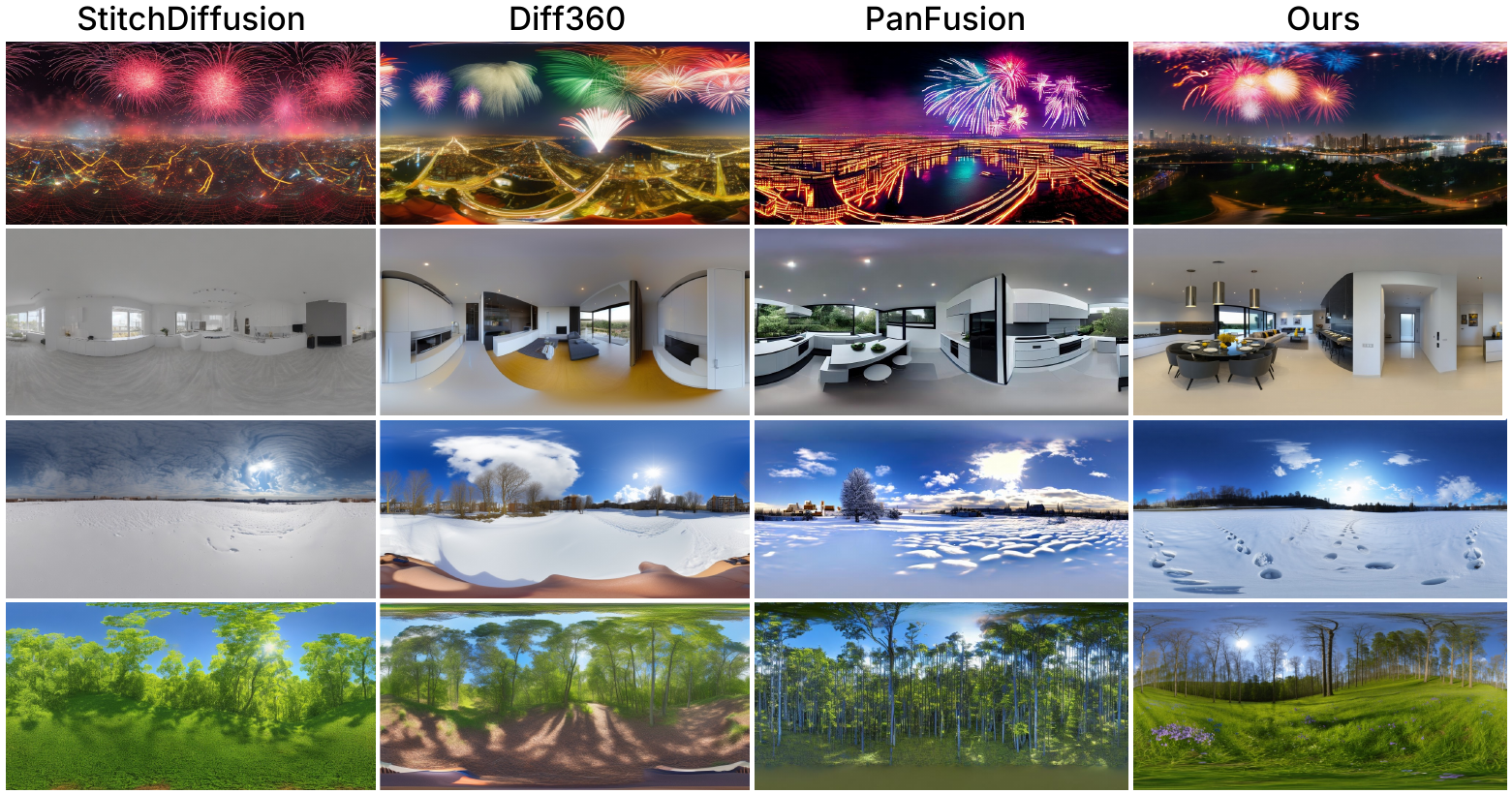}
    \caption{\textbf{Qualitative comparison of panoramic image generation methods across various scene types.} StitchDiffusion produces outputs that lack fine detail and fails on the out-of-domain fireworks example. PanFusion and Diffusion360 suffer from polar distortions and seam artifacts. In contrast, TanDiT consistently delivers detailed, globally coherent images for all prompts, including the single out-of-domain case in the final column.
    }
    \label{fig:main_results}
\end{figure}

\begin{figure}[!t]
    \centering
    \includegraphics[width=\textwidth]{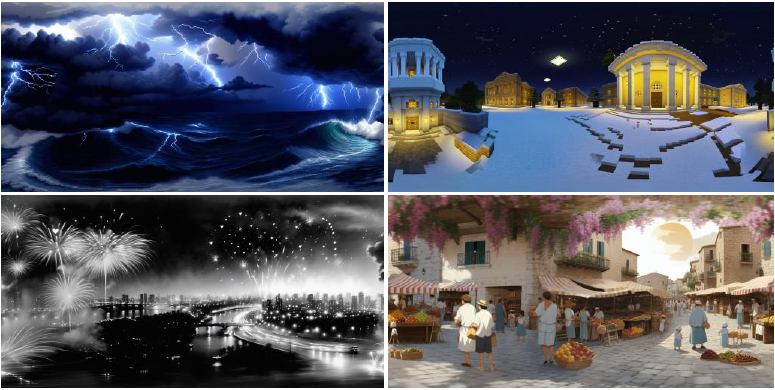}
    \caption{\textbf{Stylized panoramic images generated by TanDiT.} Examples demonstrating the model’s ability to handle a wide range of visual domains: From left to right, top to bottom: (1) traditional painting, (2) Minecraft-style rendering, (3) black-and-white charcoal sketch, and (4) Ghibli-inspired imagery. These results illustrate how TanDiT maintains spatial coherence and panoramic structure across diverse rendering styles.}
    \label{fig:stylization}
\end{figure}

Fig.~\ref{fig:main_results} and~\ref{fig:stylization} illustrate two central strengths of TanDiT, structural consistency under distortion and robust generalization to stylistically and semantically diverse prompts. 
ERP-based methods like StitchDiffusion and Diffusion360 can generate complete panoramas but frequently suffer from visual inconsistencies, especially near the poles and at the left-right seam, due to the challenges of maintaining global coherence in a highly distorted projection space. In contrast, TanDiT, by generating tangent-plane views within a structured grid and refining with ERP-aware denoising, produces globally coherent panoramas with consistent texture, lighting, and structure across the entire FoV. The fireworks example in the final column of Fig.~\ref{fig:main_results} further highlights this distinction: while other models struggle to preserve the fine-grained structure and temporal complexity of such scenes, TanDiT is able to maintain visual fidelity and spatial continuity, even in this out-of-domain setting. 

Fig.~\ref{fig:stylization} complements this with examples of stylized generation across diverse visual domains, including traditional painting, Minecraft, charcoal sketch, and Ghibli-inspired art. Without any task-specific fine-tuning, TanDiT adapts to these styles while maintaining the semantic content and spatial structure of the scene. These results show that the core design of TanDiT, which involves joint generation of tangent views and a global ERP refinement step, not only improves alignment and consistency but also enables strong generalization across prompts with varying structure and style.

\textbf{User Study.} To further evaluate our qualitative performance, we performed a 2-alternative forced-choice (2AFC) user study, with 38 participants. Each participant was shown 33 pairs of generated panoramas, rendered as a video rotating around the scene. The participants were asked to pick their preferred model based on 5 criteria, measuring the realism, consistency, and detail in the scene. We find that our model statistically outperforms all other models with $p$ < 0.05 according to the Wilson interval. Specifically, 78.1\% of users preferred our approach over Diffusion360, 79.7\% preferred ours over MultiDiffusion, 89.5\% preferred ours over StitchDiffusion, and 75.0\% preferred ours over PanFusion. Further details about our user study are included in Appendix~\ref{sec:user-study}.


\subsection{Ablations}
We perform an ablation study to assess the contributions of key components in our pipeline: patched denoising, circular padding, latent rotation, and super-resolution (Table~\ref{tab:ablations}). High-resolution panoramic generation can often introduce artifacts near image boundaries. Patched denoising addresses this by dividing the image into patches and denoising each independently, leading to improved texture quality and boundary stability. To further enhance consistency across patch borders, we pad each region using neighboring content. Additionally, we apply latent rotation during denoising to prevent overfitting to fixed spatial layouts and encourage global coherence. Disabling either mechanism results in higher discontinuity and degraded structural consistency, particularly evident in TangentFID and DS scores. While our tangent-plane design allows super-resolution at the individual view level, the ERP refinement stage must process the full image. Patched denoising enables this, even at high resolutions. Removing both super-resolution and circular padding during ERP refinement leads to a clear drop in visual quality and further increases in discontinuity and distortion metrics.

 \begin{table}[!t]
      \caption{\textbf{Ablation study evaluating the impact of key components in the TanDiT pipeline.} We report performance across standard and proposed metrics for variants of our model with specific components removed. Each ablation leads to a noticeable drop in performance, confirming the importance of patched denoising, circular padding, latent rotation, and super-resolution in maintaining global consistency, image quality, and panoramic fidelity. The best result is highlighted in \textbf{bold}, while the second-best result is indicated with an \underline{underline}.}
     \centering
     \renewcommand{\arraystretch}{1.25}
     \resizebox{\textwidth}{!}{
     \begin{tabular}{@{}l@{$\;\;$}c@{$\;\;\,$}c@{$\;\;\,$}c@{$\;\;\,$}c@{$\;\;\,$}c@{$\;\;\,$}c@{$\;\;\,$}c@{$\;\;\,$}c@{$\;\;\,$}c@{}}
            \toprule
          \textbf{Model} & \textbf{FID}$\downarrow$ & \textbf{KID}$\downarrow$ & \textbf{IS}$\uparrow$ & \textbf{CS}$\uparrow$ & \textbf{FAED}$\downarrow$ & \textbf{OmniFID}$\downarrow$ & \textbf{DS}$\downarrow$ & \textbf{TangentFID}$\downarrow$ & \textbf{TangentIS}$\uparrow$ \\
          \midrule
          TanDiT & \textbf{32.03} & \textbf{0.013} & 4.49 & \underline{23.48} & \textbf{2.56} & \textbf{49.62} & \textbf{0.0004} & \textbf{35.39} & \textbf{6.06} \\
          $\;\,$ w/o Patched Denoising & 45.50 & 0.027 & \underline{4.51} & 22.75 & \underline{2.87} & 113.80 & 0.0017 & 78.42 & 4.89 \\
          $\;\,$ w/o Circ. Pad. + Lat. Rot. & \underline{33.14} & \textbf{0.013} & 4.36 & 23.33 & \textbf{2.56} & \underline{52.24} & \underline{0.0013} & \underline{37.77} & 5.24 \\
          $\;\,$ w/o SR + Circ. Pad. & 39.26 & \underline{0.019} & \textbf{4.69} & \textbf{24.09} & 2.98 & 57.51 & 0.0021 & 40.45 & \underline{5.50} \\
          \midrule
          
     \end{tabular}}
     \label{tab:ablations}
 \end{table}

\textbf{4K Generation Results.} 
Since our model generates tangent planes as individual perspective views, existing super-resolution methods can be applied independently to each view. In our experiments, we use a 2$\times$ scale factor by default to balance speed and quality. Given the default resolution of $192 \times 192$ pixels per tangent image, applying super-resolution with a scale factor of 4$\times$ enables straightforward generation of 4K panoramas.
We provide an example of 4K output in Figure~\ref{fig:teaser}, and include further results in the Appendix.
\vspace{-8pt}
\section{Conclusion}
\vspace{-8pt}
We presented TanDiT, a diffusion-based framework for high-quality 360$^\circ$ panorama generation. In contrast to prior approaches that rely on equirectangular or cubemap projections, TanDiT operates directly on structured grids of tangent-plane images, enabling more accurate and spatially coherent synthesis. Notably, our design allows direct use of existing DiT architectures without any architectural modifications, where training proceeds on tangent-plane grids within a single diffusion loop. A final refinement step, conditioned on the intermediate equirectangular projection, further enhances global consistency and visual quality, supporting smooth transitions and arbitrary output resolutions. To better evaluate panoramic image generation, we introduced two new metrics, TangentIS and TangentFID, that capture geometric and perceptual fidelity more effectively than standard measures. Extensive experiments demonstrate that TanDiT outperforms existing baselines across multiple metrics and generalizes well to diverse styles and scene types. To support fair and standardized comparisons, we will publicly release our captioned prompt dataset and evaluation scripts alongside the proposed metrics. We expect this work to facilitate further research in 360$^\circ$ content generation.

Despite these advantages, several limitations remain. First, while the tangent-plane grid is learned using rectified optical flow, the model lacks an explicit mechanism for enforcing consistency across adjacent tangent views. As a result, minor artifacts may appear when stitching these views into an equirectangular panorama without refinement. Second, since the model is fine-tuned specifically for grid-based generation, separate weights are required for the refinement stage. Although our chosen noise level helps preserve most image details, some local changes can still occur. Lastly, as with all powerful generative models, TanDiT raises concerns about potential misuse, including the creation of misleading or synthetic content.

\section*{Acknowledgements}
Aykut Erdem acknowledges the support of the Scientific and Technological Research Council of Turkey 2247-A National Leaders Research Grant \#123C550. Hakan Çapuk, Andrew Bond, and Muhammed Burak Kızıl were supported by the AI Fellowship provided by Koç University \& Iş Bank Artificial Intelligence (KUIS AI) Research Center.


\bibliographystyle{unsrtnat}
\bibliography{references}


\appendix


\begin{center}
\section*{\centering{Appendix}}
\end{center}

\section{Contributions}
\subsection*{Core Contributors}
\begin{itemize}
    \item \textbf{Hakan Çapuk}: Trained the model, developed and tested many ideas for types of training/inference, developed the equirectangular-conditioned refinement and super-resolution methods, circular padding, running and testing of baselines, generation of results for our models and ablations, setting up image datasets, tested early captioning ideas.
    \item \textbf{Andrew Bond}: Helped with training the model and experimenting with different training losses/training methods, preparing panoramic evaluation code, developing both of the new metrics, calculation of all metrics, captioning of images, ran some ablations, and generation of style transfer results.
    \item \textbf{Muhammed Burak Kızıl}: Worked on testing our approach with Flux, running several of the baselines on our new dataset, and testing some new ideas for improving performance.
\end{itemize}
\subsection*{Partial Contributors and Advisors}
\begin{itemize}
    \item \textbf{Emir Göçen}: Helped with panoramic evaluation code for existing metrics, and set up the user study.
    \item \textbf{Erkut Erdem}: Provided technical guidance and contributed to the conceptual development of the project.
    \item \textbf{Aykut Erdem}: Provided technical supervision and served as the principal investigator overseeing the project.
\end{itemize}

\section{Methodological and Architectural Details}
\label{sec:architecture-details}

\subsection{Gnomonic Projection Formulation}
\label{sec:tangent-plane}
The gnomonic projection is a nonconformal mapping (a mapping which doesn't preserve angles) from the sphere onto the plane. We adopt the gnomonic projection to map points from the surface of a sphere to 2D image coordinates $(x, y)$ on a plane tangent to the sphere at a reference point $S$. While the projection is nonconformal, it does preserve straight lines, making it well-suited for generating perspective-consistent tangent plane views. %

More specifically, let $S$ be the point of intersection between the sphere and the projection plane. Let $\lambda_0, \phi_0$ be the central longitude and central latitude of the projection (the longitude and latitude of the point $S$). Then, for an arbitrary point with longitude and latitude $\lambda, \phi$, respectively, the resulting coordinates on the plane are given by
\begin{align}
    x &= \frac{\cos(\phi) \sin(\lambda - \lambda_0)}{\cos(c)}, \\
    y &= \frac{\cos(\phi_0)\sin(\phi) - \sin(\phi_0)\cos(\phi)\cos(\lambda - \lambda_0)}{\cos(c)},
\end{align}
where
\begin{equation}
    \cos(c) = \sin(\phi_0)\sin(\phi) + \cos(\phi_0)\cos(\phi)\cos(\lambda - \lambda_0).
\end{equation}

Here, $c$ represents the angular distance from the point $(x,y)$ to the center of the projection. Note that the gnomonic projection would map antipodal points to the same location on the sphere, so at minimum, a different plane is required for each hemisphere. The projection can be restricted even further by ensuring that any point with angular distance from the central longitude and central latitude more than half the desired FOV of each tangent plane is not projected.

To achieve full panoramic coverage while maintaining low distortion, we extract a total of 18 tangent views arranged in a fixed \(3 \times 6\) grid. The views are evenly distributed across the sphere to balance equatorial and polar regions. Overlapping fields of view between adjacent tangent planes are intentionally included to ensure smooth stitching and promote spatial continuity during training and refinement.

\subsection{Equirectangular-Conditioned Refinement}

Projecting the generated tangent-plane grid into a panoramic image via equirectangular projection often introduces visible artifacts and inconsistencies, particularly in the overlapping regions between adjacent views. These artifacts stem from limitations of the flow-matching loss used in the DiT backbone, which does not explicitly account for spatial coherence in the reprojected panorama.

To address this, we propose a refinement stage, termed \textit{Equirectangular-Conditioned Refinement}, which leverages the strong generative capacity of the pretrained DiT to improve both the visual consistency and fine details of the panorama. Importantly, prior work has shown that diffusion models are not inherently equipped to handle the geometric distortions present in equirectangular projections. Therefore, we treat the intermediate ERP image as a conditioning signal.

To mitigate the aforementioned issue, we use the previously generated intermediate equirectangular image as a conditioning input to the pretrained DiT during the refinement step. Specifically, we perturb the latent representation of the equirectangular image using noise sampled from a high timestep of the noise scheduler. This perturbation helps suppress high-frequency inconsistencies such as misalignments or localized artifact while preserving the low-frequency structure, including the global scene layout. The model is then conditioned on both this noisy latent and the original text prompt, which guides the denoising process to restore semantic alignment and visual coherence.

However, this refinement step introduces a new challenge: loop inconsistency at the horizontal boundaries of the ERP image, due to a lack of information exchange between the left and right edges. To resolve this, we implement a circular padding strategy \cite{diffusion360, panodiff}, in which the left and right edges of the latent are padded using pixel values from the opposite edges. Specifically, we pad each side with 5 columns from the opposite edge, perform denoising, and then remove the padded regions. This operation is repeated at every denoising step to maintain continuity at the image boundaries. Finally, before decoding the latent into an image, we apply circular padding once more, decode the latent, and crop the padded regions to produce a loop-consistent panorama.

\subsection{Custom Evaluation Metrics}
\label{sec:metrics_discussion}

In this section, we formally justify the design of our proposed TangentIS and TangentFID metrics, and explain their advantages over standard metrics such as Inception Score (IS), Fréchet Inception Distance (FID), and OmniFID when applied to panoramic images. 

The standard Inception Score suffers from two key issues when used on 360° images. First, the Inception network accepts 299$\times$299 inputs while equirectangular projections have a 2:1 aspect ratio. Resizing ERP images to fit the required input introduces significant distortion. Second, the Inception model is trained on perspective images, making it inherently biased toward outputs that resemble perspective-style framing such as those produced by wide-angle or NFoV-based models rather than full panoramas. To address this, TangentIS applies the Inception Score independently to each of the 18 tangent-plane images, each sized 192$\times$192, which are undistorted and perspective-consistent. This preserves spatial fidelity while maintaining compatibility with the Inception model. 

Furthermore, as discussed in the main paper, averaging the 18 IS scores corresponding to each tangent plane can unfairly reward models that perform well only in the equatorial regions. For example, wide perspective methods may generate high-quality outputs at the equator but perform poorly near the poles. A simple mean would mask these failures. To address this, TangentIS instead uses the lower bound of the 95\% confidence interval across the 18 scores. This penalizes inconsistent models and better reflects the holistic quality of 360$^{\circ}$ generation.

The same limitations apply to FID, as it also relies on the Inception model. However, we further compare our TangentFID to OmniFID, which computes FID across cubemap projections.

The cubemap representation introduces geometric distortion, especially near the edges of each cube face. This distortion increases with distance from the cube center, degrading the reliability of the extracted features. There are 3 relevant types of distortion for all representations: length distortion (the length of a line segment on the sphere vs its length on the projected image, $D_{L}(\theta)$), angular distortion (the change in angles from the sphere to the projected image, $D_{\omega}(\theta)$), and area distortion (area on the sphere vs projected area, $D_{A}(\theta)$). No perspective projection of a sphere can minimize all 3 of these distortions at once. A thorough exploration of different spherical distortions can be found in \cite{map_projections}.

Now, lets compare these distortions on our 18 tangent plane representation vs a cubemap. Table \ref{tab:distortions} shows a comparison of the different types of maximum distortion for each of the representations. We see that cubemaps always produce a higher distortion compared to the tangent plane representation, but the maximum area distortion is significantly worse for cubemaps ($2.34 \times$ higher). In particular, this means that if a cubemap face is generated as a regular perspective image without distortion, and then projected onto the sphere, there will be significant distortions near the edges of each cube face, compared to the same for our tangent plane representation. Conversely, if you want to take an existing spherical image and compute metrics with respect to that image, using a cubemap representation (like OmniFID does) introduces significantly more distortion, which can affect the accuracy of the metrics.

\begin{table}[!h]
    \caption{\textbf{A comparison of the different types of distortion introduced by our tangent plane representation compared to the cubemap representation.} Here, $D_{L}(\theta)$ is the length distortion at angle $\theta$ away from the center (the two numbers represent the distortions of a radial and tangential line, respectively), $D_{\omega}(\theta)$ is the angular distortion at angle $\theta$ away from the center, and $D_{A}(\theta)$ is the total area distortion at angle $\theta$ from the center (if we have a cube on the sphere, we assume that one set of sides is radial and the other set is tangential). We can see that the cubemap has worse distortions of every type, but the area distortion is especially bad, over twice the level of distortion as our tangent plane representation.}

    \centering
    \begin{tabular}{lccc}
    \midrule
        Representation & Max $D_{L}(\theta)$ & Max $D_{\omega}(\theta)$ & Max $D_{A}(\theta)$ \\
        \midrule
        18 Tangent Planes & 1.7/1.3 & 15.24$\degree$ & 2.22 \\
        Cubemap & 3/1.73 & 31.08$\degree$ & 5.20  \\
        \midrule
    \end{tabular}
    \label{tab:distortions}
\end{table}

Additionally, the input resolution constraint of the Inception model further limits the effectiveness of cubemap-based evaluation. With cube faces sized at $299 \times 299$, the maximum ERP resolution that can be represented without resizing is approximately $1024\times 512$. In contrast, the tangent-plane formulation allows for significantly higher representational capacity; 18 tangent views with the same resolution already cover up to roughly $1800 \times 900$ in ERP resolution. Moreover, due to the modular nature of the tangent-plane approach, even higher resolutions can be supported simply by increasing the number of tangent views. Nonetheless, to ensure fair comparison with baselines, which are all evaluated at $1024\times512$, we standardize our metrics using 18 tangent planes.

\begin{figure}[!t]
    \centering
    \includegraphics[width=\textwidth]{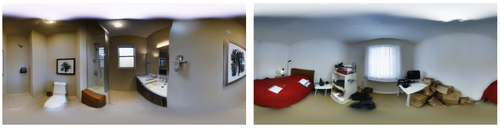}
    \caption{\textbf{Qualitative results illustrating the limitations of standard panoramic evaluation metrics}. These images are taken from our model's first-stage outputs, with no refinement stage. There are several noticeable inconsistencies, both in the equatorial and polar regions of these images. However, these images attain an OmniFID score of 45.11, quantitatively outperforming all models, including our full model with the refinement stage. This indicates that OmniFID doesn't fully correlate with panoramic image consistency and quality. Meanwhile, these images have much worse TangentFID and TangentIS scores, only better than StitchDiffusion and MultiDiffusion.}

    \label{fig:qualitative_metrics_comparison}
\end{figure}

Finally, in Figure \ref{fig:qualitative_metrics_comparison}, we provide qualitative comparisons highlighting the failure cases of standard metrics and how TangentIS and TangentFID better correlate with visual quality across the entire panoramic field of view, especially in the polar and boundary regions. Specifically, these results are taken from our first-stage model results, without any refinement. While still a well-formed panoramic image, there are notable inconsistencies in both the polar and equatorial regions of these images. However, these results outperform our own model and all baselines in OmniFID (scoring $45.11$ vs our model with refinement, which gets $49.62$). This indicates that OmniFID doesn't correlate very well with image consistency in either the polar or equatorial regions. Meanwhile, these results score poorly on both TangentFID and TangentIS ($38.73$ and $5.46$ respectively), only outperforming MultiDiffusion and StitchDiffusion (both of which achieve very poor panoramic results in general). 

\section{Implementation and Training Details}

\subsection{Dataset Composition and Preprocessing}
\label{sec:supp_dataset}

We train our model on a combined dataset composed of panoramic images sourced from Flickr360~\cite{flickr360}, Polyhaven~\cite{polyhaven}, and Matterport3D~\cite{matterport3d}. Specifically, we use 2,700 images from Flickr360, 750 from Polyhaven, and 9,000 from Matterport3D. The panoramas in Flickr360 are provided at a resolution of 1024$\times$2048, while those from Polyhaven are 2048$\times$4096. Matterport3D offers panoramic scenes in cubemap format, which we convert into 1024$\times$2048 equirectangular images as a preprocessing step.

Following the approach of~\cite{omnifusion}, we extract 18 tangent-plane views from each equirectangular panorama. Due to the distortion properties of equirectangular projection, tangent planes extracted from polar regions exhibit greater distortion and cover a larger area in the scene. To account for this, we extract fewer tangent views from these regions. Specifically, we divide the equirectangular image into four horizontal rows: the first and fourth rows correspond to the north and south poles, from which we extract 3 tangent planes each; the second and third rows are closer to the equator, and we extract 6 tangent planes from each. This results in a total of 18 tangent views, distributed non-uniformly to better match the spatial characteristics of the projection. Each view is downscaled to \(192 \times 192\), and arranged into a \(3 \times 6\) grid, yielding a final grid image of size \(576 \times 1152\), which serves as input to our model during training. 

In our experiments, we aim to arrange the tangent planes in a grid such that neighboring views are placed adjacent to one another, approximating their spatial relationships in the equirectangular projection. However, due to the way we extract 18 tangent planes, a perfect spatial correspondence between grid positions and their original layout on the panorama is not possible.
To achieve the closest possible alignment, we construct the grid in such a way that the first row contains the six tangent planes sampled from the polar regions (three from the north pole and three from the south pole). The second and third rows consist of the twelve tangent planes extracted from the equatorial region (six per row). This arrangement ensures that the central rows of the grid—containing the majority of the scene—preserve correct local adjacency, which helps the model capture spatial continuity across neighboring views.

A key challenge in panoramic generation is the lack of publicly available datasets with accompanying text captions. To overcome this, we generate dense descriptions using the LLaVA-One-Vision model~\cite{llavaonevision}. However, these captions often exceed the context window of the text encoders used by many baseline models. To ensure fair comparison, we use LLaMA 2~\cite{llama2} to produce concise summaries, making the captions compatible with varying encoder capacities.

The prompt we provided to LLaVA-One-Vision is

\fbox{
\begin{minipage}{\textwidth}
\textsf{\small Give a detailed caption of the following equirectangular projection of a panoramic image. Be detailed about all of the important entities, textures, and colors in the different parts of the scene. Provide enough detail that a text-to-image diffusion model would be able to reconstruct the scene.}
\end{minipage}
}

and the prompt provided to LLaMA 2 is

\fbox{
\begin{minipage}{\textwidth}
\textsf{\small Given a long text prompt that describes a panoramic scene, \{longer\_caption\}, summarize this text prompt to a shorter one that describes what the whole scene looks like without losing important details.}
\end{minipage}
}

\subsection{Model Configurations}

\textbf{DiT Backbone.}
We adopt Stable Diffusion 3.5 Large (SD3)~\cite{sd3} as the backbone for our diffusion transformer (DiT) architecture. SD3 incorporates three text encoders—two based on CLIP and one based on a T5 encoder—to support both short and long captions. Input latents are first divided into non-overlapping patches, which are then embedded with positional encodings and processed jointly with the corresponding text embeddings.

Unlike Stable Diffusion 1 and 2, which rely on a U-Net-based noise prediction network, SD3 replaces this component with a transformer-based architecture consisting of 37 Multimodal DiT (MM-DiT) blocks. Each MM-DiT block contains dual processing streams for image latents and text embeddings, enabling the integration of visual and textual modalities. These blocks employ Layer Normalization, Multi-Head Attention, and Feed-Forward layers for expressive and efficient information flow. After the final MM-DiT block, the latent features are projected through a linear layer, unpatchified, and used to predict the denoised output. SD 3.5-Large uses a hidden size of 2432, and a patch size of $2\times2$.

\textbf{Super Resolution Backbone.} 
In our experiments, we use VarSR \cite{varsr} as the super resolution model in our panorama generation pipeline. VARSR's pipeline begins by encoding the low-resolution input into prefix tokens that condition the generation process across multiple scales. To preserve spatial structure, VARSR introduces Scale-Aligned Rotary Positional Encoding (SA-RoPE), aligning tokens spatially across scales within an autoregressive transformer. A lightweight diffusion refiner is employed to estimate quantization residuals and recover high-frequency details lost during tokenization. Finally, an image-based classifier-free guidance (CFG) mechanism leverages quality-aware embeddings to guide the generation toward perceptually superior results without additional training.

\subsection{Training Protocols}

We train our model on a combined dataset of 12,450 grid-caption samples, as described in Section~\ref{sec:supp_dataset}, using a batch size of 8. To fine-tune the pre-trained Stable Diffusion 3.5-Large model~\cite{sd3}, we adopt the LoRA method~\cite{lora}, leveraging the AdamW optimizer~\cite{adamw} with a learning rate of $1 \times 10^{-4}$ and a constant learning rate scheduler over 10 epochs. Training is conducted on a single Nvidia A40 GPU and takes approximately 40 GPU-hours in total, using the standard rectified flow loss.

\subsection{Inference Protocols}
\begin{figure}[!t]
    \centering
    \includegraphics[width=\linewidth]{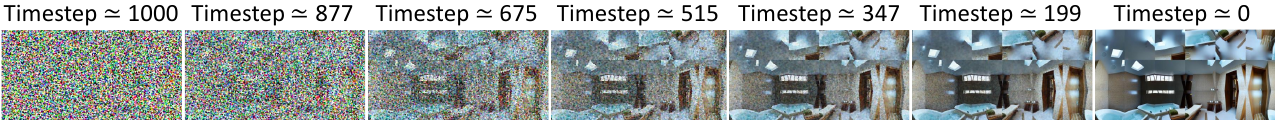}
    \caption{\textbf{Generated panorama by timestep during inference of TanDiT.} TanDit maps random noise to a grid of tangent planes conditioned on a given text.}
    \label{fig:noise_figure}
\end{figure}

In the first stage of our panorama generation pipeline, we use the fine-tuned TanDiT model to generate a grid of 18 tangent planes conditioned on a given panorama caption. We employ a diffusion process with 28 timesteps and set the guidance scale to 7.0. To allow for longer captions, we increase the \texttt{max\_sequence\_length} to 512. Inference process of TanDiT is visualized in Figure~\ref{fig:noise_figure}.

In the second stage of Equirectangular-Conditioned Refinement, we perturb the intermediate panorama latent with noise corresponding to a specific timestep. This noisy latent, along with the original caption, is passed to the pre-trained Stable Diffusion 3 model. The model then performs denoising over the specified number of timesteps to produce the final refined panorama.

\section{Runtime Analysis}

TanDiT generates all tangent-plane views simultaneously using a structured grid layout, making the runtime of this stage primarily dependent on the underlying diffusion model. In our experiments, we use Stable Diffusion 3.5 Large~\cite{sd3} with 28 inference steps, resulting in a tangent grid generation time of approximately 20 seconds.

Next, the grid is split into 18 individual tangent-plane images, each of which is processed by a pretrained super-resolution model. By default, we use VarSR~\cite{varsr} with a 2\( \times \) upscale factor. The complete super-resolution step for all tangent planes takes approximately 25 seconds.

In the final stage, the upscaled tangent views are reprojected into an equirectangular panorama and encoded into a latent. This latent is then perturbed with noise at a mid-range diffusion timestep and refined using the pretrained DiT model. Our default setting uses 17 denoising steps; however, due to the patched denoising strategy where each timestep involves two passes to maintain spatial consistency, this refinement stage takes approximately 30 seconds.

In total, our default pipeline (2\( \times \) upscaling with patched refinement) completes in about 75 seconds per image. When super-resolution is disabled (thus enabling single-pass denoising), the runtime drops to approximately 45 seconds. Conversely, increasing the upscale factor to 4\( \times \), which requires 8 patch passes due to increased resolution, raises the total runtime to around 140 seconds.

\section{Ablation Studies}

\subsection{Ablating the Usage of Equirectangular Conditioned Refinement}

\begin{figure}[!h]
    \centering
    \begin{subfigure}[b]{0.49\textwidth}
        \includegraphics[width=\linewidth]{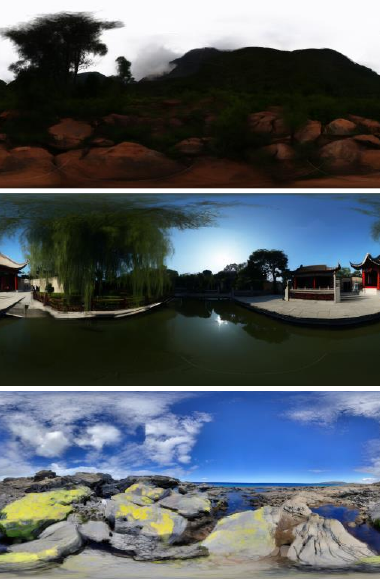}
        \caption{Before Refinement}
        \label{fig:sub1}
    \end{subfigure}
    \hfill
    \begin{subfigure}[b]{0.49\textwidth}
        \includegraphics[width=\linewidth]{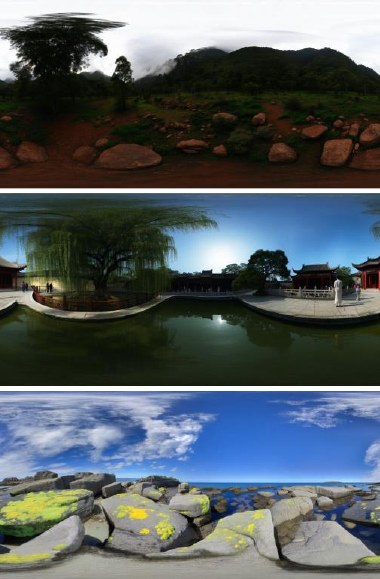}
        \caption{After Refinement}
        \label{fig:patched_denoising}
    \end{subfigure}
    \caption{\textbf{Example panoramas before and after applying the Equirectangular Conditioned Refinement step.} Refinement step mitigates the artifacts and inconsistencies arising from projecting the tangent planes to a panoramic image, and enables the generation of a visually coherent panorama.}
    \label{fig:refinement_figure}
\end{figure}

We further demonstrate the effectiveness of our Equirectangular-Conditioned Refinement strategy by comparing panoramas generated before and after its application. As shown in Figure~\ref{fig:refinement_figure}, this refinement step mitigates artifacts and inconsistencies introduced by projecting overlapping tangent planes into the panoramic view. Furthermore, the strong generative capabilities of the pretrained DiT model enhances fine details in the scene, enabling the generation of visually coherent and higher-quality panoramic images.

\subsection{Ablating the Usage of Patched Denoising in Refinement Stage}

In our refinement stage, we observe that the underlying DiT model struggles to denoise the upscaled equirectangular panorama at a resolution of \(1024 \times 2048\). This leads to noticeable textural degradation, particularly near the polar regions, while the equatorial regions exhibit higher visual quality. We hypothesize that this issue arises because most diffusion transformer models, such as Stable Diffusion 3~\cite{sd3} and Flux~\cite{flux}, are predominantly trained on square-shaped images.

\begin{figure}[!h]
    \centering
    \begin{subfigure}[b]{0.49\textwidth}
        \includegraphics[width=\linewidth]{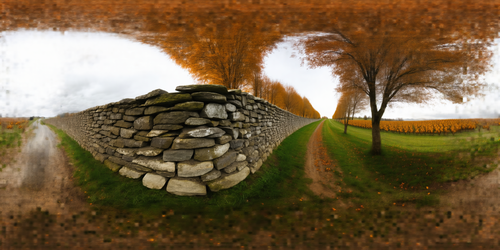}
        \caption{Without Patched Denoising}
        \label{fig:sub1}
    \end{subfigure}
    \hfill
    \begin{subfigure}[b]{0.49\textwidth}
        \includegraphics[width=\linewidth]{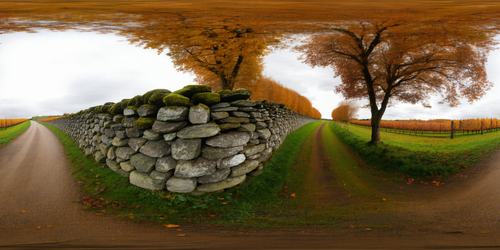}
        \caption{With Patched Denoising}
        \label{fig:patched_denoising}
    \end{subfigure}
    \caption{\textbf{Effect of patched denoising on high-resolution panoramic generation.} (a) Without patched denoising, the model struggles to maintain texture quality and consistency across the image, resulting in visible artifacts and degradation near the boundaries. (b) With patched denoising, the image retains sharper details, improved coherence, and better visual quality across the entire panoramic scene, demonstrating the effectiveness of this strategy in high-resolution settings.}
    \label{fig:patch_denoising}
\end{figure}

To address this limitation, we divide the panorama into non-overlapping \(1024 \times 1024\) square patches, allowing the pretrained DiT model to operate more effectively while still generating rectangular panoramas with a \(1{:}2\) aspect ratio. To maintain scene coherence across patch boundaries, we apply circular padding to each patch using pixels from neighboring regions at every denoising step. A qualitative comparison of panoramas refined with and without the patched denoising strategy is provided in Figure~\ref{fig:patch_denoising}. Moreover, this strategy facilitates information flow between adjacent patches and enables high-quality generation at higher resolutions, such as 4K (\(4096 \times 2048\)), as shown in Figure~\ref{fig:4k}.

\subsection{Ablating Latent Rotation and Circular Padding in Patched Denoising}

Our patched denoising strategy inherently introduces potential inconsistencies between patches, as we operate on non-overlapping square regions independently. To promote information sharing across neighboring patches, we apply circular padding in each denoising step and again before VAE decoding. Specifically, each patch is padded using pixels from its adjacent patches, allowing the model to denoise patches separately while maintaining global scene coherence.

\begin{figure}[!t]
    \centering
    \begin{subfigure}[b]{0.49\textwidth}
        \includegraphics[width=\linewidth]{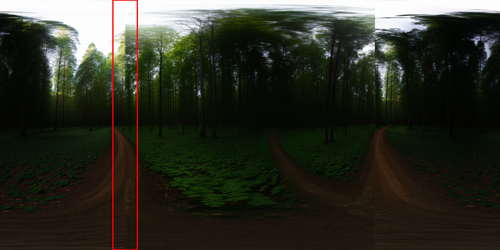}
        \caption{Without Latent Rotation and Circular Padding}
        \label{fig:sub1}
    \end{subfigure}
    \hfill
    \begin{subfigure}[b]{0.49\textwidth}
        \includegraphics[width=\linewidth]{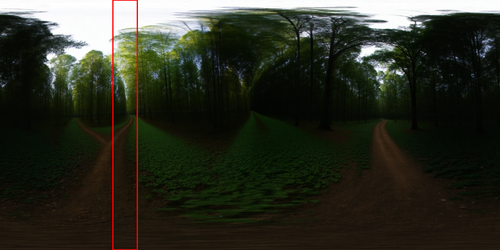}
        \caption{With Latent Rotation and Circular Padding}
        \label{fig:sub2}
    \end{subfigure}
    \caption{\textbf{Effect of latent rotation and circular padding in patched denoising.} Images are horizontally rotated by 90° to better visualize the loop-consistency. Original left-right loop regions are marked with red. (a) Without circular padding, there is no sufficient information flow between the left and right edges of the image during the denoising. Thus, the model is prone to generating panoramic images that are not loop-consistent. (b) With circular padding combined with latent rotation, the model is able to denoise the patches while keeping the whole scene visually coherent.}
    \label{fig:circularpaddingandlatentrotation}
\end{figure}

However, we observe that this approach can still introduce seams in the interior regions of the patches. This occurs because the model primarily focuses on aligning the boundaries between patches, while potentially neglecting internal consistency within each patch. To address this, we adopt a latent rotation strategy: during each denoising step, we cyclically rotate the entire latent representation along the horizontal axis. This encourages the model to attend to different spatial regions throughout the denoising process, rather than focusing on a fixed area. After denoising, we reverse the rotation to restore the original spatial alignment and ensure that the generated panorama remains consistent with the input prompt in terms of content layout and position. The combined effect of circular padding and latent rotation is illustrated in Figure~\ref{fig:circularpaddingandlatentrotation}.

\subsection{Ablating the Noise Level Used in Refinement Stage}

To justify our choice of $T_{\text{high}} \simeq 800$ which corresponds to $17$ steps of inference, we conduct ablation experiments with two alternative settings: \( T_{\text{high}} \simeq 700 \) (13 steps) and \( T_{\text{high}} \simeq 900 \) (21 steps). The results are presented in Table \ref{tab:refinement_levels}. We observe that using a higher timestep, $T_{\text{high}} \simeq 900$, improves generic image quality metrics such as Inception Score and CLIP Score, likely due to the model having more flexibility to reshape fine details. However, this comes at the cost of degrading panorama-specific metrics (FAED, OmniFID, DS, TangentFID, TangentIS), as more of the original scene structure is lost during denoising. Conversely, using a lower timestep, $T_{\text{high}} \simeq 700$, better preserves the original panoramic structure, resulting in higher scores on metrics tailored to global consistency. That said, it also leaves more residual artifacts and inconsistencies due to insufficient refinement, reflected in reduced IS and CLIP scores.

\begin{figure}[!t]
    \centering
    \begin{subfigure}[b]{0.49\textwidth}
     \includegraphics[width=\linewidth]{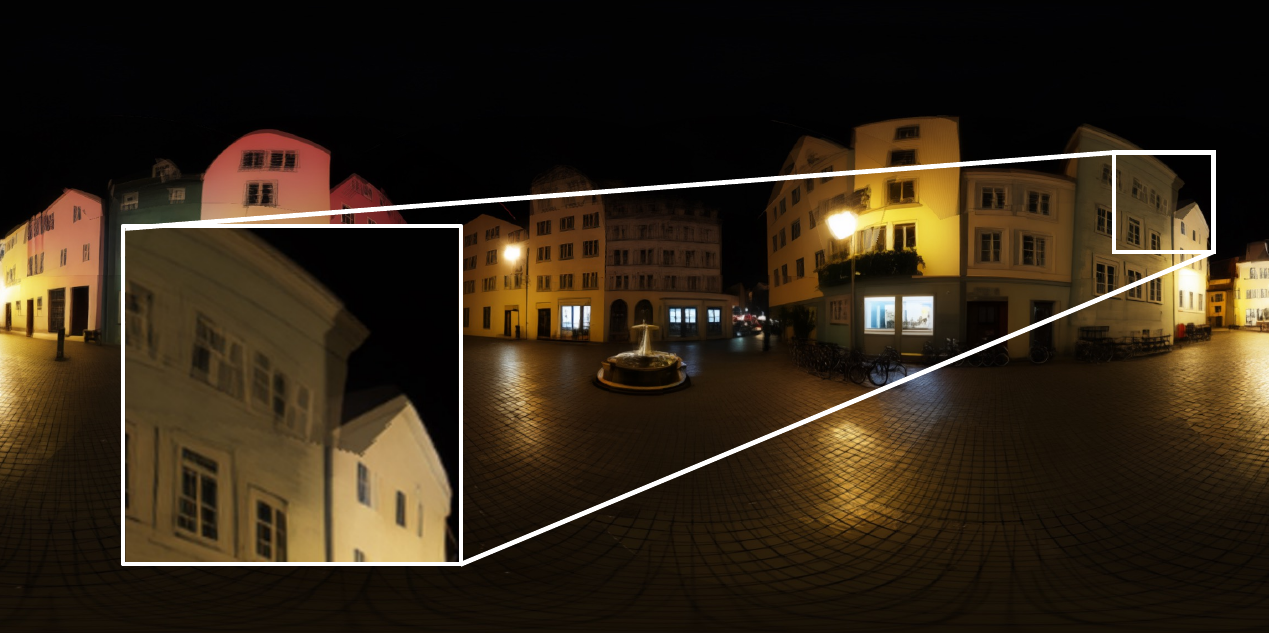}
        \caption{Initial Generation Result}
        \label{fig:sub1}
    \end{subfigure}
    \hfill
    \begin{subfigure}[b]{0.49\textwidth}
    \includegraphics[width=\linewidth]{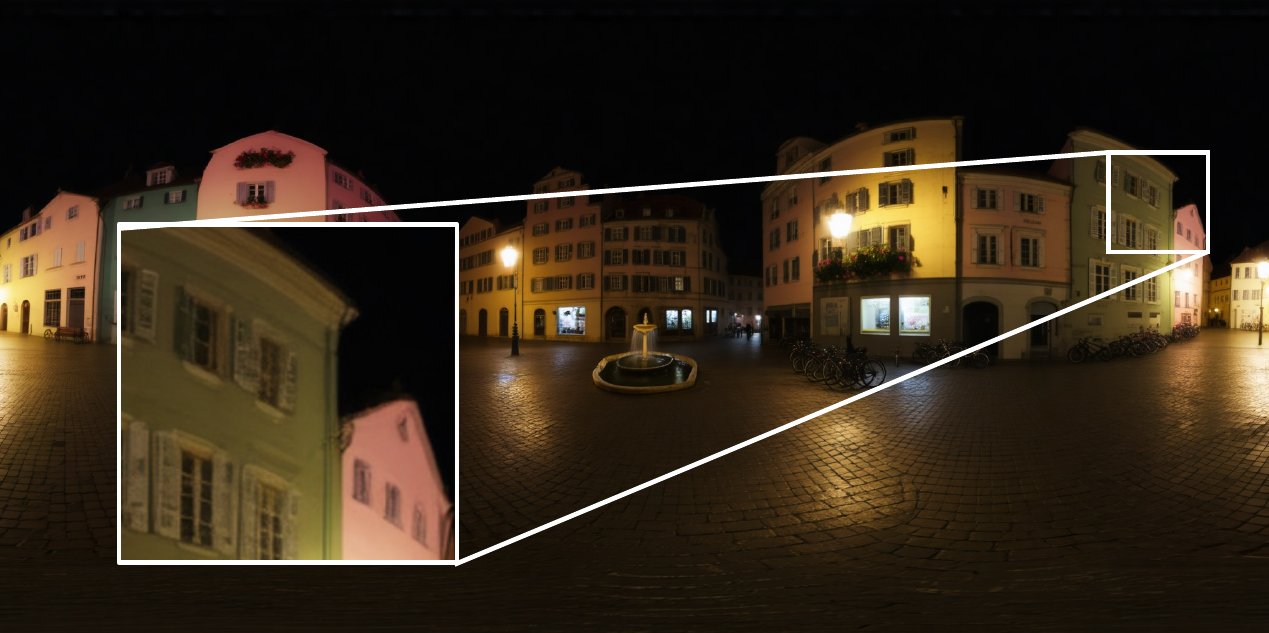}
        \caption{Refinement with $T\simeq$ 700}
        \label{fig:sub2}
        \end{subfigure}
    \begin{subfigure}[b]{0.49\textwidth}
     \includegraphics[width=\linewidth]{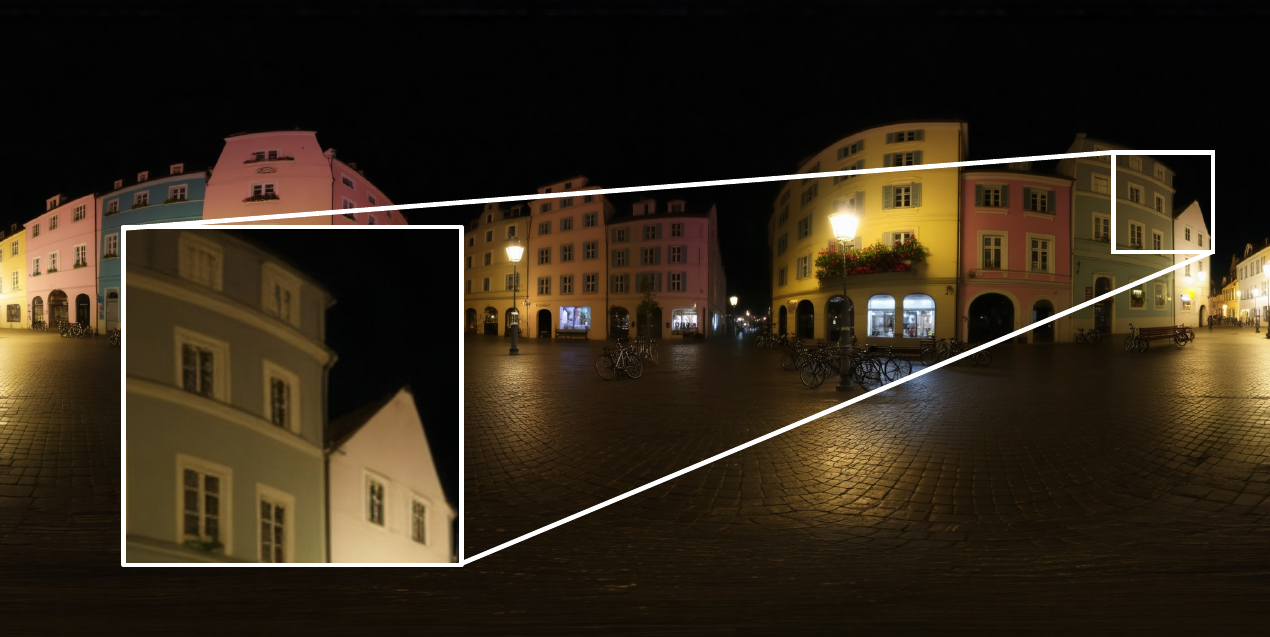}
        \caption{Refinement with $T\simeq$ 800}
        \label{fig:sub1}
    \end{subfigure}
    \hfill
    \begin{subfigure}[b]{0.49\textwidth}
    \includegraphics[width=\linewidth]{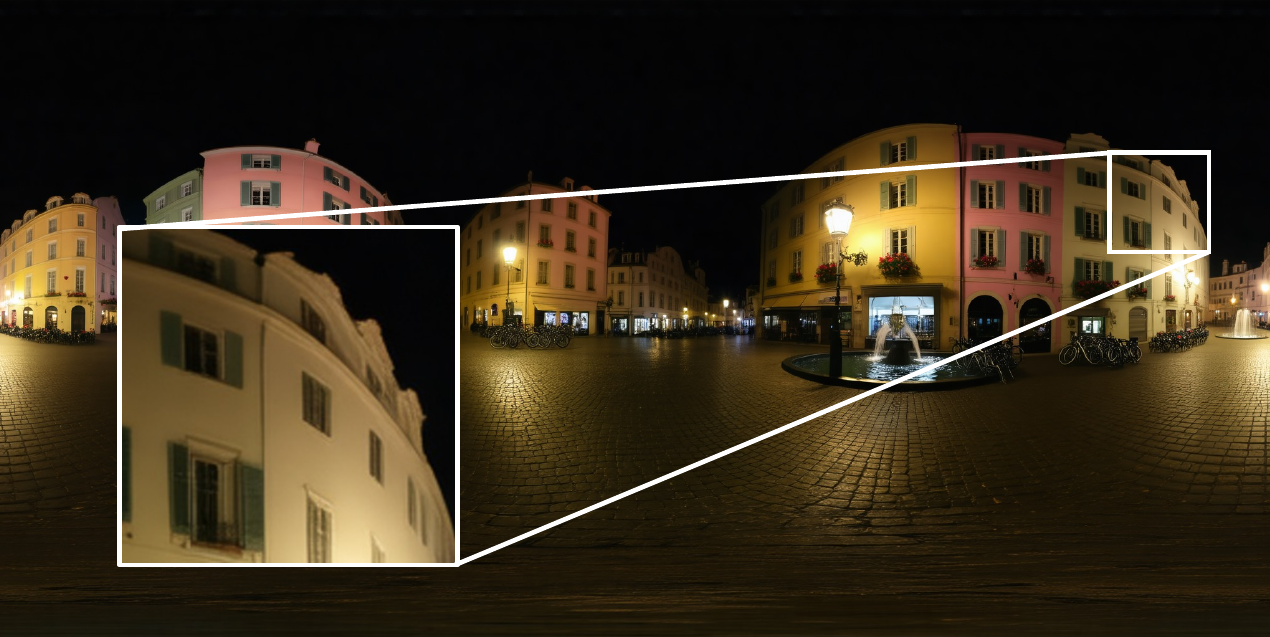}
        \caption{Refinement with $T\simeq$ 900}
        \label{fig:sub2}
    \end{subfigure}
    \caption{\textbf{Effect of timestep in the Equirectangular-Conditioned Refinement process.} We visualize the effect of different noise levels used during refinement in TanDiT. Starting from the initial equirectangular panorama, we apply refinement using noise injected at various timesteps ($\simeq$700, $\simeq$800, $\simeq$900). Lower timesteps yield subtle improvements, primarily reducing high-frequency artifacts, while higher timesteps introduce more aggressive changes that may alter fine-grained content. As shown in the insets, refinement improves visual coherence and detail consistency in overlapping regions. In our final model, we adopt a timestep of $\simeq$800 as a balance between visual fidelity and effective artifact removal.}
    \label{fig:refinement_stepsize}
\end{figure}

This trade-off is visually illustrated in Figure~\ref{fig:refinement_stepsize}, which shows the output of our refinement process using each of the tested timestep settings. Lower timesteps primarily reduce high-frequency noise, while higher timesteps yield stronger visual changes that may alter prompt-relevant details. Our default choice of $T_{\text{high}} \simeq 800$ allows for a balance between these two extremes, preserving essential panoramic content while allowing sufficient refinement to improve visual quality and semantic alignment.

\begin{table}[!t]
    \caption{Comparison of different timesteps for our equirectangular-conditioned refinement process. Picking the number of timesteps for refinement is crucial for balancing the quality of the refinement, while still preserving the crucial structure of the panoramic image. Going too low, ($700$), preserves more structure but fails to refine the images enough, so the quality is lower. Meanwhile, going too high, $(900)$, destroys a lot of the panoramic structure, but allows for higher-quality images. A timestep of ($800$) allows for balancing these two to obtain superior results.}

    \centering
    \resizebox{\textwidth}{!}{
    \begin{tabular}{l@{$\;\;$}c@{$\;\;$}c@{$\;\;$}c@{$\;\;$}c@{$\;\;$}c@{$\;\;$}c@{$\;\;$}c@{$\;\;$}c@{$\;\;$}c}
        \midrule
        \textbf{Num Timesteps} & \textbf{FID} & \textbf{KID} & \textbf{IS} & \textbf{CS} & \textbf{FAED} & \textbf{OmniFID} & \textbf{DS} & \textbf{TangentFID} & \textbf{TangentIS} \\
        \midrule
         Ours ($\simeq 800$) & \textbf{32.03} & \textbf{0.013} & \underline{4.49} & \underline{23.48} & \underline{2.56} & \underline{49.62} & \textbf{0.0004} & \textbf{35.39} & \textbf{6.06} \\ 
         ($\simeq 700$) & \underline{32.33} & \textbf{0.013} & 4.31 & 23.13 & \textbf{2.55} & \textbf{49.49} & \textbf{0.0004} & \underline{36.80} & \underline{5.65} \\ 
         ($\simeq 900$) & 37.72 & \underline{0.015} & \textbf{4.80} & \textbf{24.06} & 2.64 & 61.38 & 0.0006 & 41.37  & 5.60 \\
         \midrule
    \end{tabular}}
    \label{tab:refinement_levels}
\end{table}

\subsection{Pretrained Stable Diffusion with Our Captions}
To validate that the performance gains achieved by our method stem from both the proposed training procedure and post-processing steps, and are not merely a result of using a stronger backbone model (Stable Diffusion 3), we conduct a control experiment. Specifically, we provide the same captions to the default Stable Diffusion 3 model, without any LoRA fine-tuning or refinement steps, and evaluate its outputs using the same metrics. To further assist the baseline model, we modify the caption with the phrase "\textsf{\small A panoramic image of ...}". The results are summarized in Table~\ref{tab:ours_vs_pretrained_sd3}. As shown, the outputs from the unmodified baseline are significantly worse across all key metrics (with the exception of Inception and CLIP scores, whose limitations we have already discussed in the main paper).

\begin{table}[!t]
    \caption{\textbf{Comparison between our proposed method (TanDiT) and the pretrained SD3 model.} We evaluate the default SD3 model without fine-tuning or post-processing, with and without the trigger phrase "A panoramic image of ...". TanDiT outperforms both variants across all key panoramic-specific metrics, confirming that our gains are due to the proposed training and refinement pipeline, not just the underlying backbone. Although SD3 achieves higher IS and CLIP scores, these are less reliable for panoramic evaluation, as discussed in the main paper.}

    \centering
    \resizebox{\textwidth}{!}{
    \begin{tabular}{l@{$\;\;$}c@{$\;\;$}c@{$\;\;$}c@{$\;\;$}c@{$\;\;$}c@{$\;\;$}c@{$\;\;$}c@{$\;\;$}c@{$\;\;$}c}
        \midrule
        \textbf{Model} & \textbf{FID} & \textbf{KID} & \textbf{IS} & \textbf{CS} & \textbf{FAED} & \textbf{OmniFID} & \textbf{DS} & \textbf{TangentFID} & \textbf{TangentIS} \\
        \midrule
         TanDiT & \textbf{32.03} & \textbf{0.013} & 4.49 & 23.48 & \textbf{2.56} & \textbf{49.62} & \textbf{0.0004} & \textbf{35.39} & \textbf{6.06} \\ 
         Pretrained SD3 & 67.63 & 0.041 & \textbf{6.29} & \textbf{25.89} & \underline{4.29} & 79.13 & 0.0050 & 63.16 & 2.50 \\ 
         $+$ Trigger Phrase & \underline{48.12} & \underline{0.023} & \underline{5.09} & \underline{24.96} & 4.63 & \underline{69.97} & \underline{0.0039} & \underline{51.64}  & \underline{2.55} \\
         \midrule
    \end{tabular}}
    \label{tab:ours_vs_pretrained_sd3}
\end{table}

\subsection{Ablating the Impact of Our Captions on the Results}
Stable Diffusion 3 can handle substantially longer captions than earlier models, up to 512 tokens, compared to just 77 in Stable Diffusion 1 and 2. As discussed previously, we used summarized captions for training the baseline models and for computing the CLIP score, to ensure compatibility with their shorter context lengths. To verify that our model’s improved performance is not simply due to access to longer, more detailed captions, we conduct two additional experiments: (1) we train our proposed method (TanDiT) using the summarized captions, and (2) we train all baseline models using the full, detailed captions (noting that these will be truncated when exceeding each model’s token limit). The results, presented in Table~\ref{tab:caption_effect_results}, confirm that the performance gap cannot be attributed solely to caption length.

\begin{table}[!t]
    \caption{\textbf{Effect of caption length on model performance.} We evaluate the impact of long vs. summarized captions on both TanDiT and the baselines. TanDiT is trained using summarized captions, while the baselines are trained with full, detailed captions (truncated based on each model’s token limit). Despite this reversal in caption advantage, TanDiT continues to outperform all baselines across all panoramic-specific metrics, confirming that its superior performance is not solely due to access to longer text inputs.}
    \centering
    \resizebox{\textwidth}{!}{
    \begin{tabular}{l@{$\;\;$}c@{$\;\;$}c@{$\;\;$}c@{$\;\;$}c@{$\;\;$}c@{$\;\;$}c@{$\;\;$}c@{$\;\;$}c@{$\;\;$}c}
        \midrule
        \textbf{Model} & \textbf{FID} & \textbf{KID} & \textbf{IS} & \textbf{CS} & \textbf{FAED} & \textbf{OmniFID} & \textbf{DS} & \textbf{TangentFID} & \textbf{TangentIS} \\
        \midrule
         TanDiT & 32.03 & 0.013 & 4.49 & 23.48 & 2.56 & 49.62 & 0.0004 & 35.39 & 6.06 \\ 
         Panfusion & 34.60 & 0.023 & 4.71 & 20.65 & 3.55 & 57.86 & 0.0005 & 47.01 & 5.97 \\
         StitchDiffusion & 77.53 & 0.053 & 4.74 & 19.78 & 9.90 & 115.29 & 0.0015 & 69.23 & 2.45 \\
         Diffusion360 & 48.42 & 0.028 & 4.04 & 22.11 & 4.37 & 55.79 & 0.0006 & 46.29 & 2.48 \\
         MultiDiffusion & 73.16 & 0.044 & 7.27 & 23.95 & 4.31 & 87.78 & 0.0032 & 61.76 & 2.42 \\
         \midrule
    \end{tabular}}
    \label{tab:caption_effect_results}
\end{table}

\subsection{Ablating Our Ordering of the Grid}
\begin{figure}[!t]
  \centering
  \begin{tabular}{cc}
    \begin{subfigure}[b]{0.45\textwidth}
      \includegraphics[width=\linewidth]{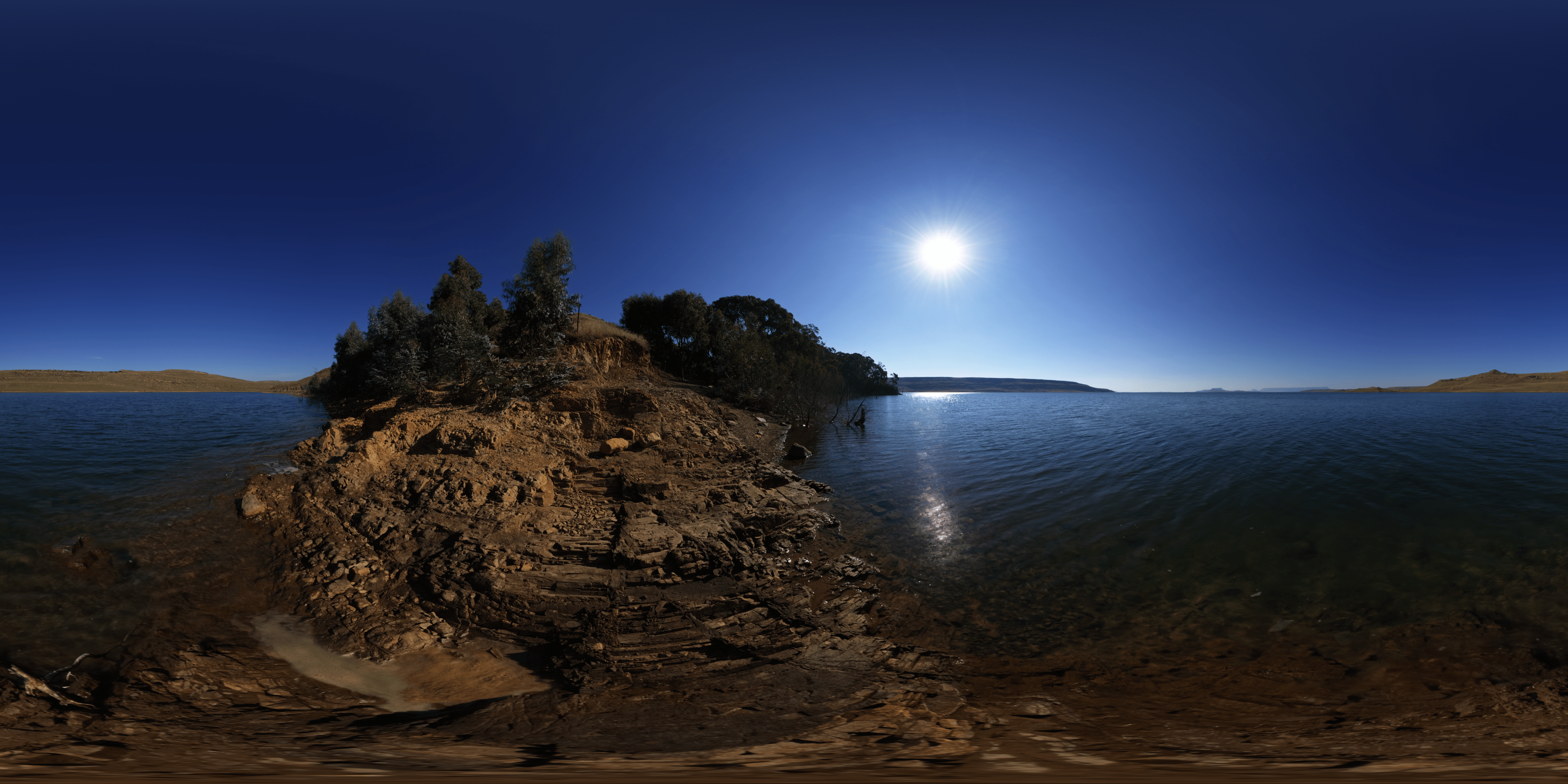}
      \caption{Equirectangular Projection}
    \end{subfigure} &
    \begin{subfigure}[b]{0.45\textwidth}
      \includegraphics[width=\linewidth]{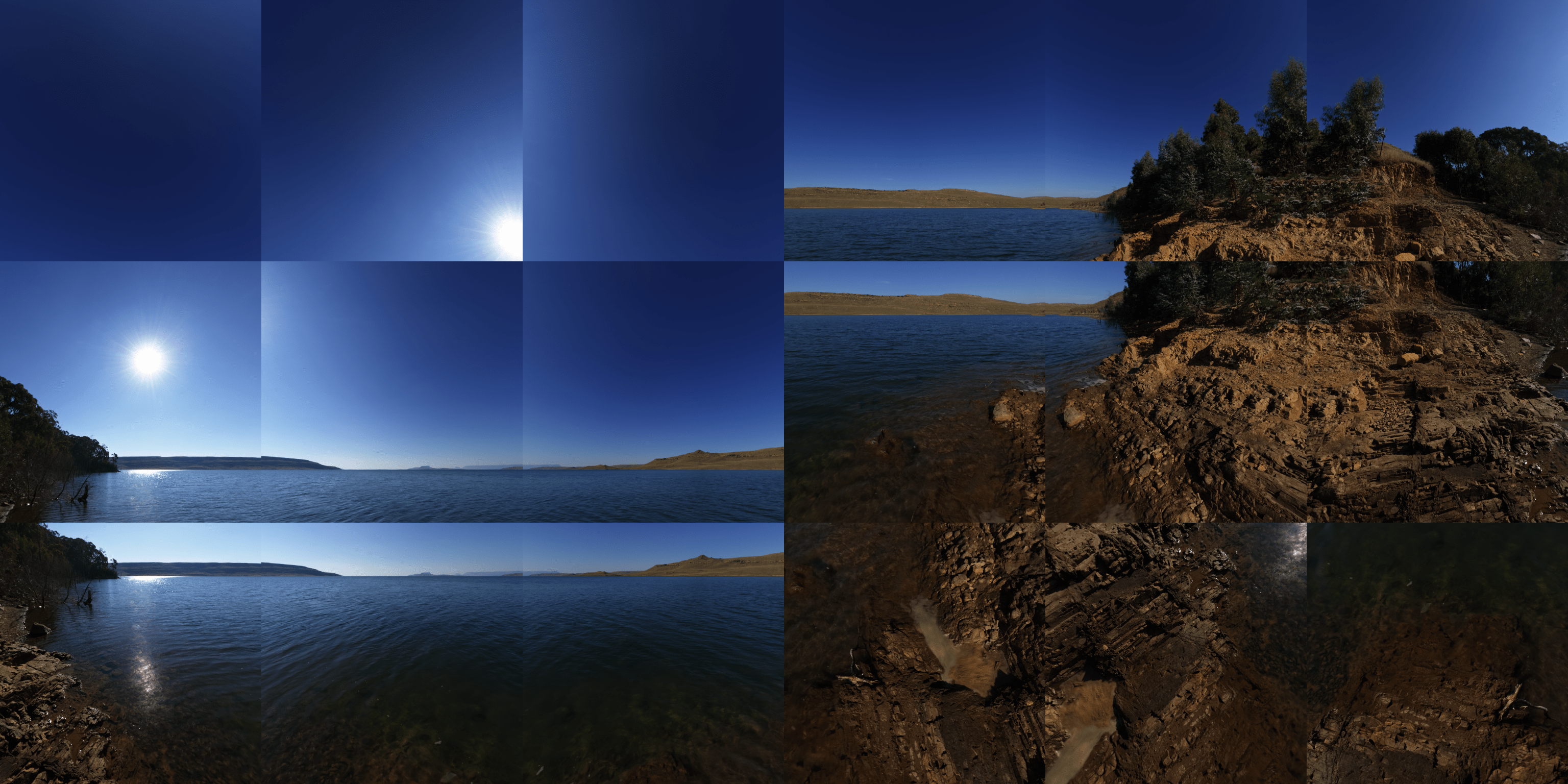}
      \caption{First version - Row Ordered}
    \end{subfigure} \\
    \begin{subfigure}[b]{0.45\textwidth}
      \includegraphics[width=\linewidth]{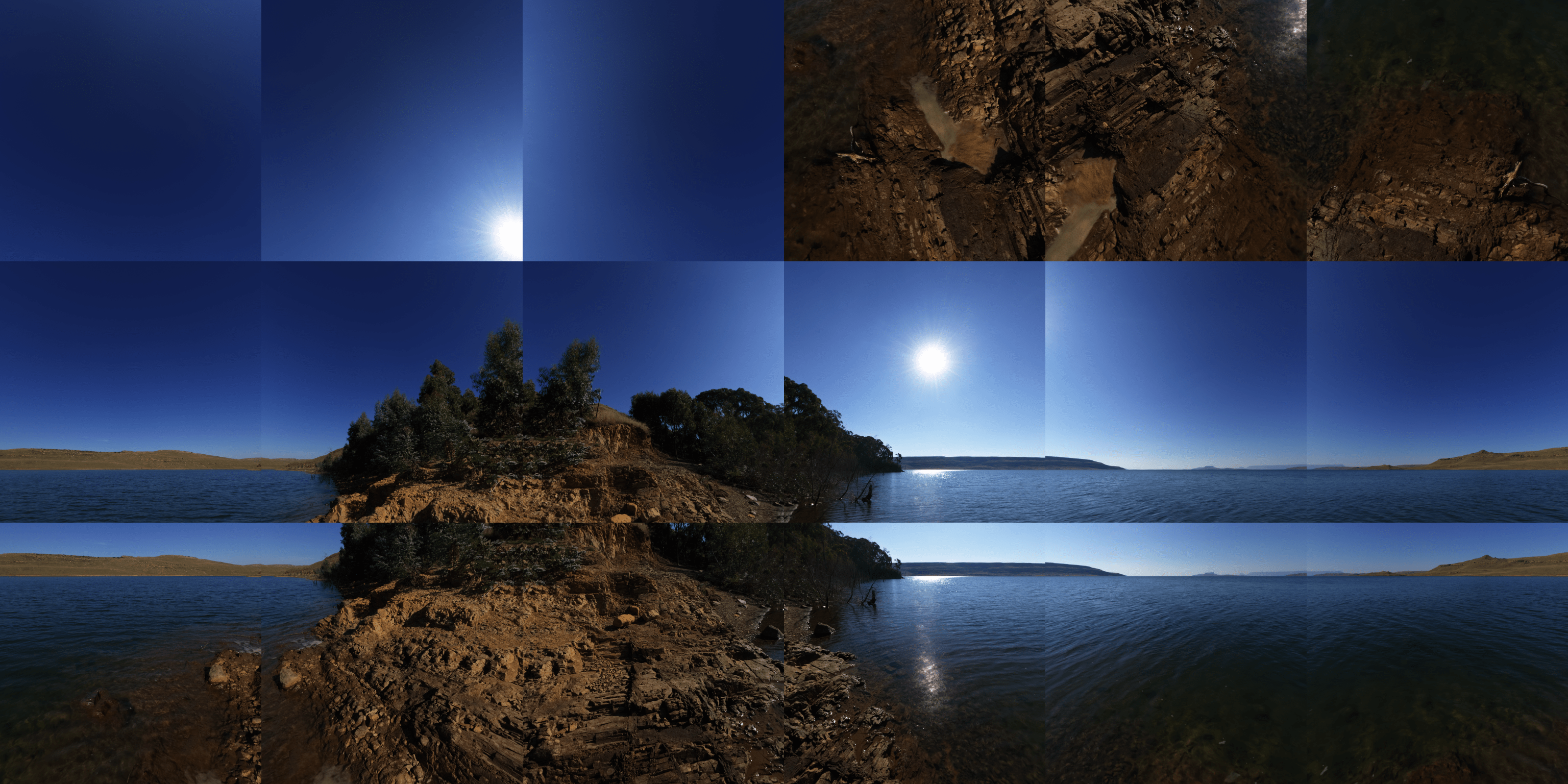}
      \caption{Second version - Our custom ordering}
    \end{subfigure} &
    \begin{subfigure}[b]{0.45\textwidth}
      \includegraphics[width=\linewidth]{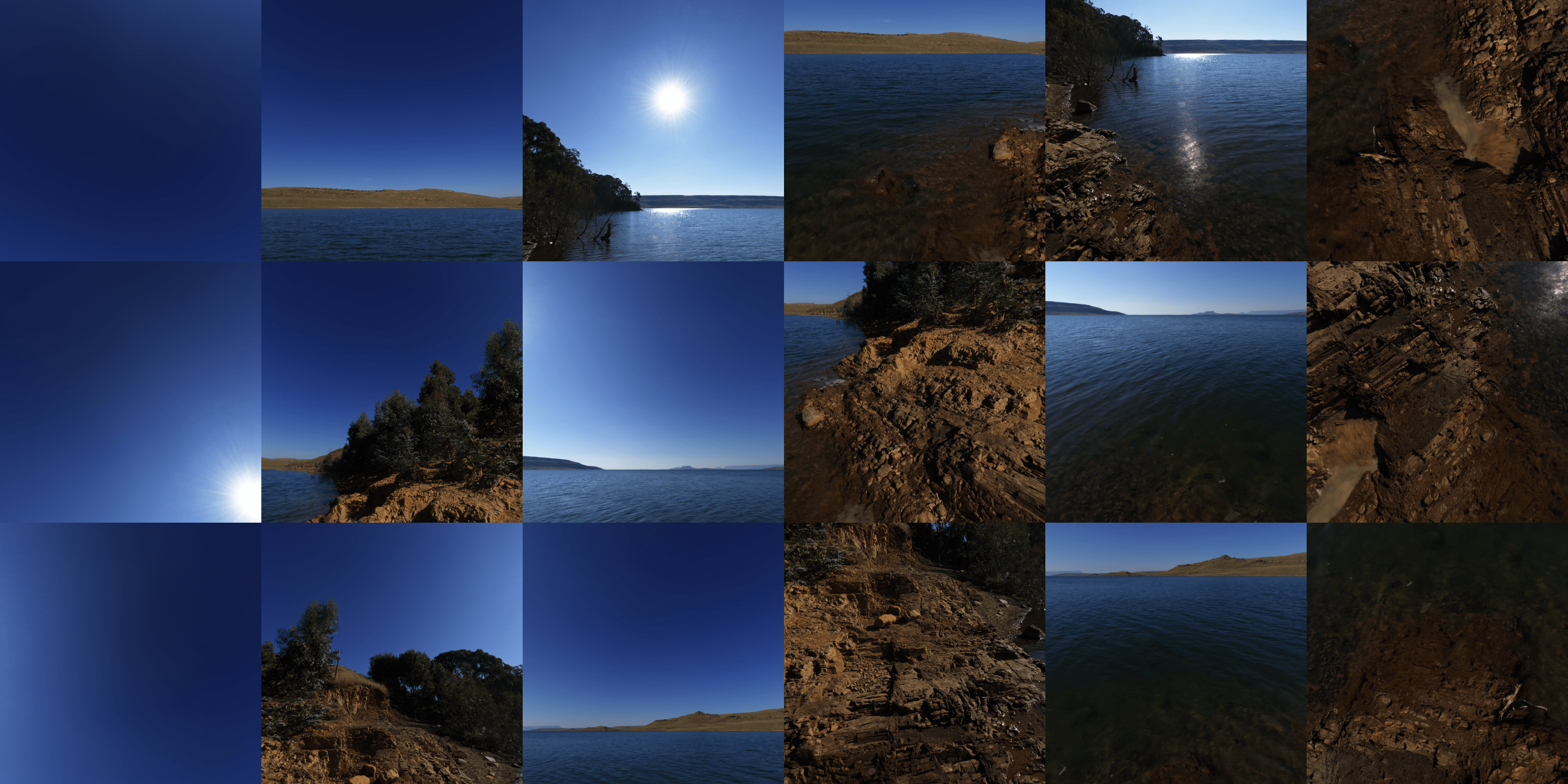}
      \caption{Third version - Column Ordered}
    \end{subfigure}
  \end{tabular}
  \caption{ \textbf{A comparison of panorama representations.} (a) Tangent planes are arranged row-wise from top to bottom. (b) Tangent planes from the bottom row of the equirectangular projection are moved to the top row of the tangent grid. This adjustment preserves the spatial order of the middle (equatorial) regions—typically the most informative parts of the scene. (c) Tangent planes are arranged column-wise from left to right.}
  \label{fig:grid_orders}
\end{figure}

While we adopt a specific grid ordering as shown in Figure~\ref{fig:grid_orders}(c), there exist numerous alternative arrangements that can maintain a degree of local consistency between neighboring tangent planes.

We evaluate our chosen configuration against two alternatives: (1) a row-wise ordering, where tangent planes are arranged left-to-right, top-to-bottom without adjustment. This would correspond to relocating the three top-right tangent planes to the bottom (Figure~\ref{fig:grid_orders}(b)), and (2) a column-wise ordering, proceeding top-to-bottom, left-to-right (Figure~\ref{fig:grid_orders}(d)).

For perspective image-based metrics (FID, KID, IS, CS), all three configurations yield comparable results. However, our proposed ordering achieves superior performance on panorama-specific metrics (OmniFID, TangentFID, TangentIS), which motivated its selection. Quantitative results comparing these grid arrangements are presented in Table~\ref{tab:grid_ordering}.

\begin{table}[!t]
    \centering
    \caption{Comparison of different orderings for our grid layout. We can see that the perspective image metrics are all very similar, while our layout does better than the rest on the panoramic image metrics.}
    \begin{tabular}{l@{$\;\;$}c@{$\;\;$}c@{$\;\;$}c@{$\;\;$}c@{$\;\;$}c@{$\;\;$}c@{$\;\;$}c@{$\;\;$}c@{$\;\;$}c}
        \midrule
        \textbf{Model} & \textbf{FID} & \textbf{KID} & \textbf{IS} & \textbf{CS} & \textbf{FAED} & \textbf{OmniFID} & \textbf{DS} & \textbf{TangentFID} & \textbf{TangentIS} \\
        \midrule
        Ours & \underline{32.03} & \underline{0.013} & \underline{4.49} & 23.48 & \underline{2.56} & \textbf{49.62} & 0.0004 & \textbf{35.39} & \textbf{6.06} \\
        Row-wise & 32.40 & \textbf{0.012} & \textbf{4.52} & \underline{23.55} & 4.03 & \underline{52.91} & 0.0004 & 36.58 & \underline{5.47} \\
        Column-wise & \textbf{31.74} & \textbf{0.012} & 4.30 & \textbf{23.63} & \textbf{1.67} & 54.36 & 0.0004 & \underline{36.41} & 5.40\\
        \midrule
    \end{tabular}
    \label{tab:grid_ordering}
\end{table}

\subsection{Ablating Number of Tangent Planes in the Grid}

To evaluate the effect of the number of tangent planes on model performance, we experiment with generating 10 and 26 tangent planes per scene, in addition to our default setting of 18. The results of the metrics are shown in Table \ref{tab:num_tangent_planes}. In the case of 26 tangent planes, we observe that the underlying DiT model struggles to learn the grid structure effectively, likely due to the increased complexity introduced by a denser layout. In contrast, both the 10 and 18 image configurations result in successful grid learning and the generation of visually coherent panoramas.

However, using a higher number of tangent planes offers advantages for high-resolution panorama generation. Since each tangent plane can be independently passed through a super-resolution pipeline, increasing their count allows the final equirectangular panorama to achieve higher resolution without requiring aggressive upscaling. This reduces the risk of introducing artifacts, as more native-resolution tangent views are available to cover the output equirectangular projected space.

\begin{table}[!t]
    \centering
    \caption{\textbf{Comparison of our model with varying levels of tangent planes in the grid.} We see that our choice of 18 tangent planes outperforms both other choices in the panoramic-specific metrics. Using 10 planes does lead to higher distortion per plane, affecting the final panoramic results. Meanwhile, the model struggles to properly learn the grid structure in the case of 26 tangent planes, leading to suboptimal results. }
    \begin{tabular}{l@{$\;\;$}c@{$\;\;$}c@{$\;\;$}c@{$\;\;$}c@{$\;\;$}c@{$\;\;$}c@{$\;\;$}c@{$\;\;$}c@{$\;\;$}c}
        \midrule
        \textbf{Num Planes} & \textbf{FID} & \textbf{KID} & \textbf{IS} & \textbf{CS} & \textbf{FAED} & \textbf{OmniFID} & \textbf{DS} & \textbf{TangentFID} & \textbf{TangentIS} \\
        \midrule
        Ours (18) & \textbf{32.03} & \underline{0.013} & 4.49 & \underline{23.48} & \underline{2.56} & \textbf{49.62} & \textbf{0.0004} & \textbf{35.39} & \textbf{6.06} \\
        10 & 32.04 & \textbf{0.012} & \underline{4.50} & \textbf{23.52} & \textbf{1.78} & \underline{53.12} & \underline{0.0005} & \underline{36.10} & \underline{5.53} \\
        26 & 42.81 & 0.019 & \textbf{4.61} & 23.01 & 6.34 & 77.58 & \underline{0.0005} & 47.64 & 5.04\\
        \midrule
    \end{tabular}
    \label{tab:num_tangent_planes}
\end{table}

\subsection{Ablating the Use of Super-Resolution}

\begin{figure}[!t]
    \centering
    \includegraphics[width=1.0\textwidth]{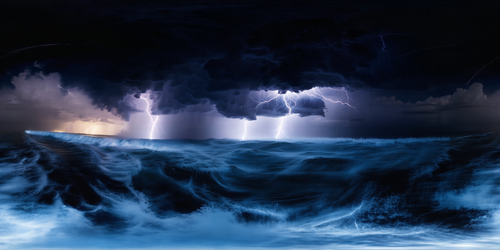}
    \caption{\textbf{A 4k image generated by TanDiT.} Employing the proposed patched denoising technique in the Equirectangular-Conditioned Refinement step, we are able to generate 4k images.}
    \label{fig:4k}
\end{figure}

Figure~\ref{fig:4k} demonstrates that TanDiT can effectively generate high-quality panoramas at resolutions up to 4K. To quantify the impact of super-resolution in achieving this, we perform an ablation comparing results with and without applying super-resolution to the generated tangent-plane images. As shown in Table~\ref{tab:using_superresolution}, applying super-resolution consistently improves performance across most metrics, especially those sensitive to texture and detail (e.g., FID, TangentFID, FAED). By employing our default 2$\times$ super-resolution strategy, TanDiT enhances spatial fidelity within each tangent plane prior to stitching and refinement. Additionally, integrating super-resolution with patched denoising in the Equirectangular-Conditioned Refinement stage provides flexible control over output resolutions, such as producing detailed 2K or 4K panoramas.

\begin{table}[!t]
    \centering
    \caption{\textbf{Comparison of our model’s performance metrics with and without super-resolution}. Applying super-resolution (default: 2$\times$) to each tangent plane leads to improvements across several metrics. This enhancement stems from the small native size of each tangent plane (192$\times$192), which may otherwise struggle to capture fine scene details.}
    \resizebox{\textwidth}{!}{
    \begin{tabular}{l@{$\;\;$}c@{$\;\;$}c@{$\;\;$}c@{$\;\;$}c@{$\;\;$}c@{$\;\;$}c@{$\;\;$}c@{$\;\;$}c@{$\;\;$}c}
        \midrule
        \textbf{Super-Resolution?} & \textbf{FID} & \textbf{KID} & \textbf{IS} & \textbf{CS} & \textbf{FAED} & \textbf{OmniFID} & \textbf{DS} & \textbf{TangentFID} & \textbf{TangentIS} \\
        \midrule
        Ours (Yes) & \textbf{32.03} & \textbf{0.013} & 4.49 & 23.48 & \textbf{2.56} & \textbf{49.62} & \textbf{0.0004} & \textbf{35.39} & \textbf{6.06} \\
        No & 38.61 & 0.018 & \textbf{4.66} & \textbf{24.11} & 2.97 & 59.35 & 0.0007 & 41.97 & 5.69 \\
        \midrule
    \end{tabular}}
    \label{tab:using_superresolution}
\end{table}

\section{Additional Qualitative Results}

\subsection{Generating Out-of-Domain Stylized Panoramas with TanDiT}
Leveraging the powerful generative capabilities of the underlying DiT model within the Equirectangular-Conditioned Refinement stage, TanDiT is capable of synthesizing high-quality stylized panoramas well beyond its original training distribution. By conditioning on descriptive textual prompts, TanDiT successfully produces panoramas exhibiting diverse artistic styles, such as "Pixel Art", "Watercolor Painting" and  "Monochrome". This flexibility enables users to explore a wide variety of visually distinct aesthetics, significantly broadening TanDiT’s potential applications, from digital art creation to immersive media content.

\begin{figure}[!t]
    \centering
    \includegraphics[width=\linewidth]{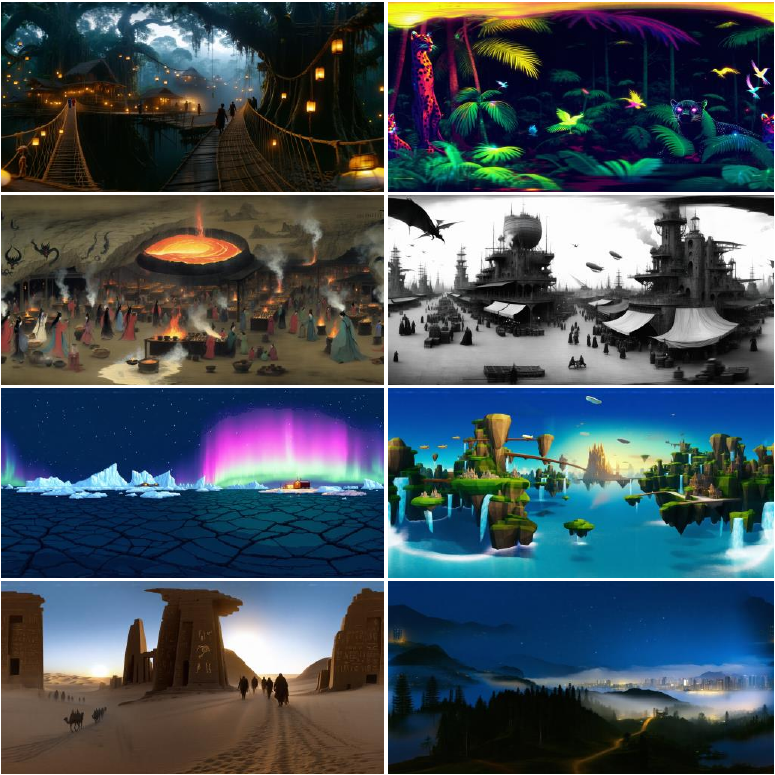}
    \caption{\textbf{Out-of-domain stylized panoramas generated by TanDiT.} TanDiT, combined with the Equirectangular-Conditioned Refinement method is able to generalize to many out-of-domain styles.}
    \label{fig:more_styles}
\end{figure}

Figure~\ref{fig:more_styles} provides additional visual examples demonstrating TanDiT's ability to generalize across these out-of-domain styles. Notably, these results illustrate that TanDiT maintains spatial coherence and panoramic consistency even when generating highly abstract or stylistically complex imagery. Future work could further investigate the integration of style-specific fine-tuning or few-shot learning to enhance quality and control in specialized stylistic domains.

    \label{fig:grid_construction}

\subsection{Qualitative Comparison with CubeDiff}
CubeDiff~\cite{cubediff} is a panoramic outpainting model that takes a perspective image and a text caption as input and generates a panoramic output by producing a cubemap, with the input perspective image used as one of its faces. Although the code is not publicly available, we provide qualitative comparisons based on the results presented on their project webpage, highlighting how our method avoids the consistency issues observed in CubeDiff.

\begin{figure}[!t]
    \centering
    \includegraphics[width=0.8\linewidth]{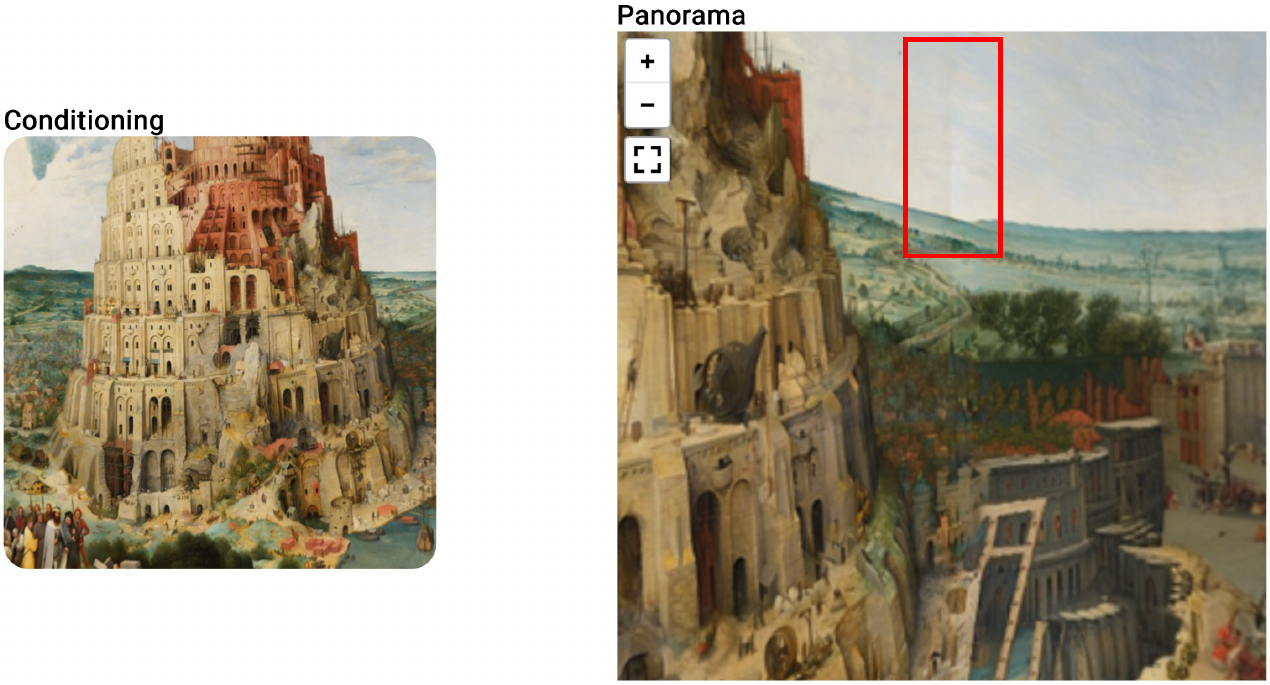}
    \caption{\textbf{Example of inconsistency in CubeDiff’s panorama generation.} The left image shows the perspective input used to condition CubeDiff. The right image is the resulting equirectangular panorama, where a visible seam appears near the center due to misalignment and distortion from cubemap face warping. This illustrates a common failure mode in CubeDiff, where treating each cubemap face as an undistorted perspective view leads to artifacts at face boundaries. Best viewed zoomed-in.}
    \label{fig:cubediff_example_1}
\end{figure}

CubeDiff generates each cubemap face independently, applying modified attention layers to enable limited interaction between the faces. The final panorama is obtained by converting the cubemap into an equirectangular projection. While this attention mechanism introduces some amount of coherence over cube faces, it does not fully resolve inconsistencies at the seams. Notably, distortions inherent to the cubemap projection, stronger near the face boundaries than at the center, are not appropriately modeled. Since CubeDiff treats each face as an undistorted perspective image, this results in visual artifacts when the faces are stitched together, particularly in regions with high distortion near the cube edges. See Section \ref{sec:metrics_discussion} for a further discussion about the distortion produced by cubemaps near the edges. For example, Figure~\ref{fig:cubediff_example_1} shows a prominent seam at the center of the image, caused by misalignment and warping during cubemap conversion. Although this is a particularly clear case, similar issues can be seen in many other results on their website. In some indoor scenes, seams may be harder to detect due to symmetric layouts (e.g., four-wall rooms), but inconsistencies still persist.

\section{Loop-Consistency Analysis}

\begin{figure}[!t]
    \centering
    \includegraphics[width=0.94\linewidth]{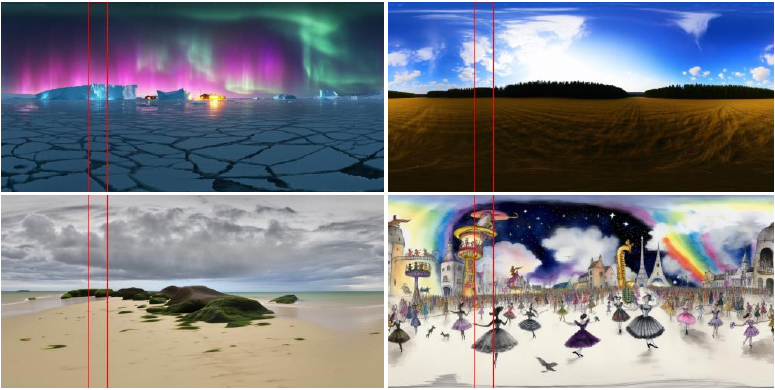}
    \caption{\textbf{Loop Consistency in the refined panoramas.} Refined panorama images are rotated by 90 degrees horizontally and marked with a red rectangle to qualitatively show the left-right continuity. Utilizing circular padding in the refinement step allows for a sufficient information flow in the left and right edges, thus enabling the generation of a fully loop-consistent panorama.}
    \label{fig:loop_consistency}
\end{figure}

To support our claim that the generated panoramas exhibit strong loop-consistency, we provide qualitative visualizations under various settings. For improved clarity, each panorama is rotated by 90 degrees, allowing the left and right boundaries to be shown side by side. The boundary regions, where seamless continuity is most critical, are highlighted with red rectangles in Figure~\ref{fig:loop_consistency}. These visualizations demonstrate that our refinement strategy, combined with circular padding, enables smooth transitions across the horizontal edges of the panorama.

\section{User Study Details}
\label{sec:user-study}
As described in Section 5.2 of the main paper, we conducted a user study via Qualtrics to visually assess the effectiveness of our method. Participants were presented with paired comparisons between our model’s output and that of a baseline model, and asked to select the image they preferred based on a given prompt. A total of 38 volunteers participated in the study and were provided with the following evaluation criteria:

\fbox{\begin{minipage}{\textwidth}
\textsf{\scriptsize You'll review panoramic images generated from text prompts. Please rate each image based on the following:}

\textsf{\scriptsize \textbf{Relevance}: Does the image faithfully represent the prompt? Are key elements included?}

\textsf{\scriptsize \textbf{Realism}: Does it look natural and believable? Are there any distortions or artifacts, especially near seams or poles?}

\textsf{\scriptsize \textbf{Seam Consistency}: Are the left and right edges smoothly connected? Is spatial consistency maintained?}

\textsf{\scriptsize \textbf{Coverage}: Is the scene immersive and complete, without blank areas or awkward repetitions?}

\textsf{\scriptsize {\textbf{Overall Impression}: How would you rate the image’s quality as a whole?}}
\end{minipage}
}

\begin{figure}[!t]
    \centering
    \fbox{
 \includegraphics[width=0.85\linewidth]{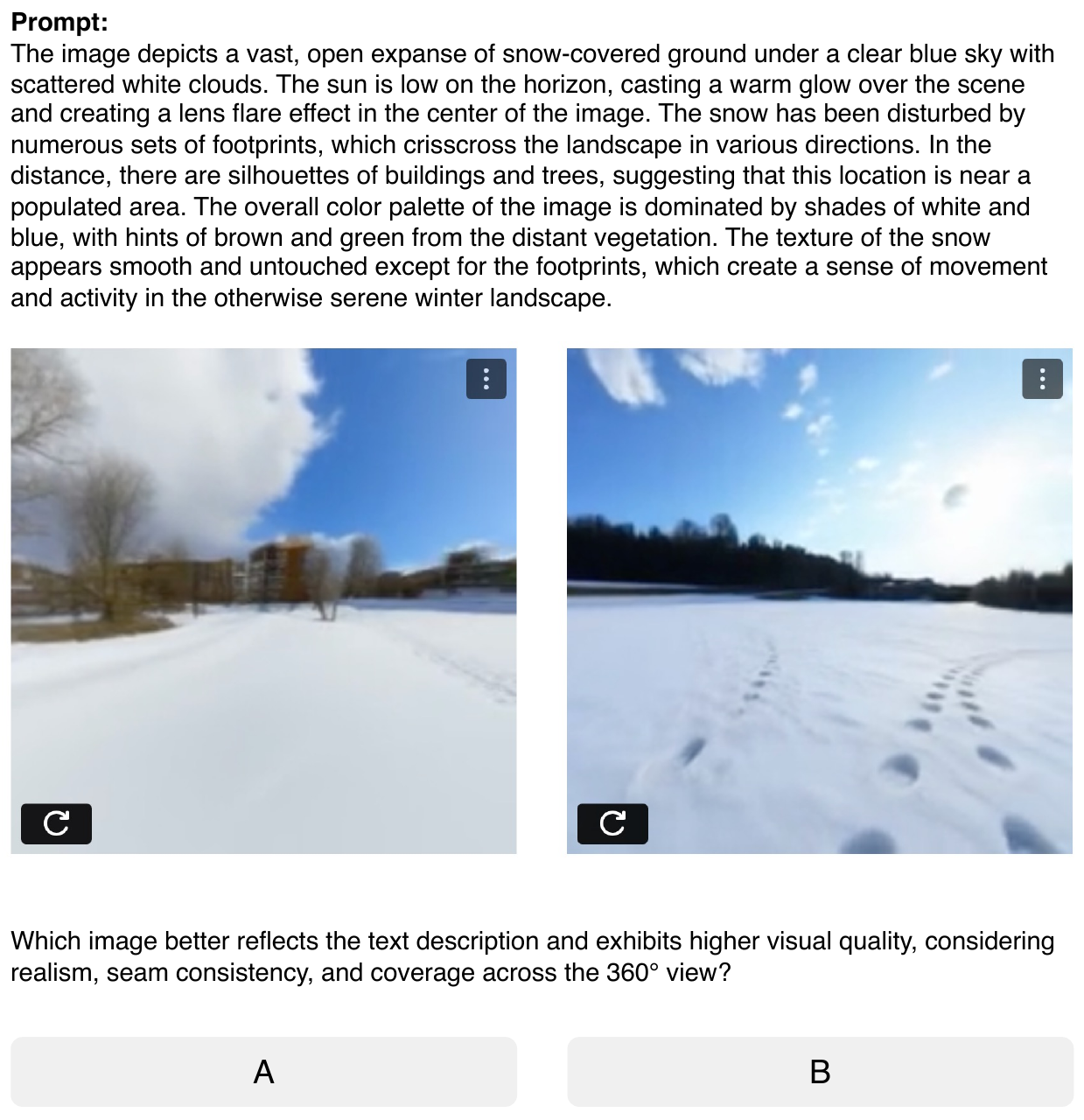}}
    \caption{\textbf{Screenshot of the survey layout.} The participants are asked to choose the best generation result with high relevance to the text prompt, realism, seam consistency, coverage, and overall impression.
    }
    \label{fig:userstudy_example}
\end{figure}

Figure~\ref{fig:userstudy_example} shows an example question from the user interface. Participants compared two videos rendered from panoramic images with identical camera settings, resolution, and rotation speed. For each question, one image was generated by our method and the other by a randomly selected baseline. The left/right positions were randomized to avoid positional bias.

\begin{figure}[!t]
  \begin{center}   \includegraphics[width=0.7\linewidth]{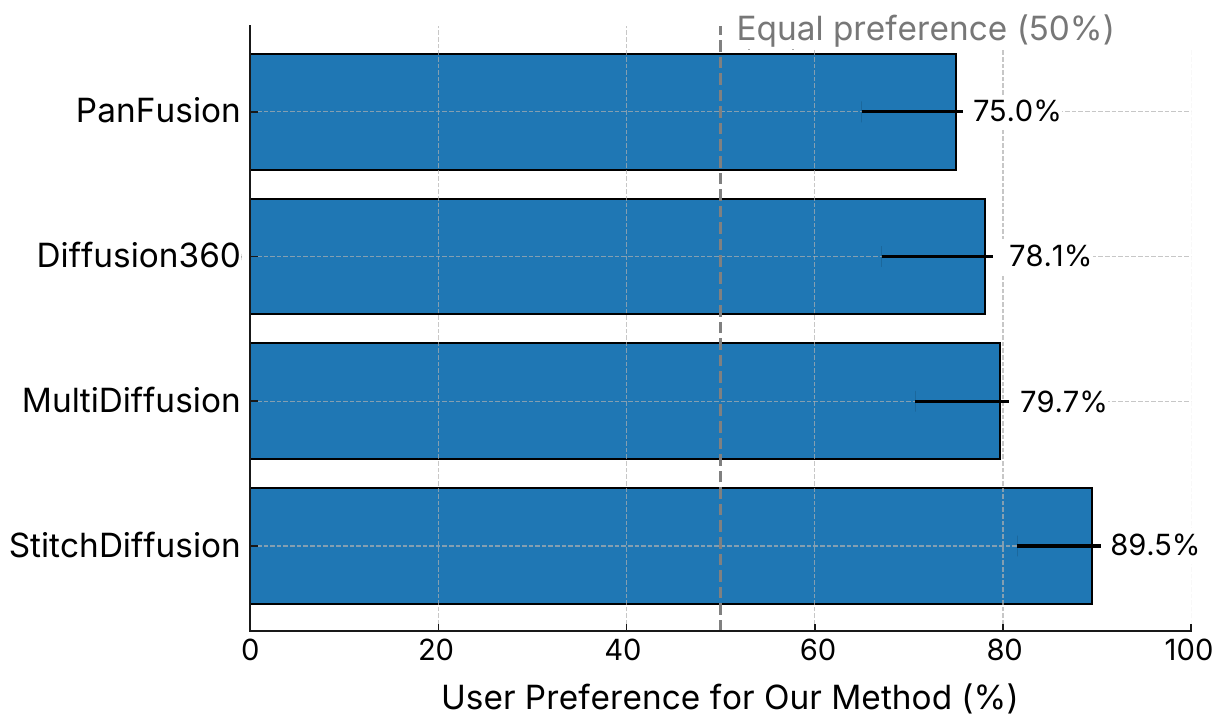}
  \end{center}
  \caption{\textbf{Results of the pairwise user preference study comparing our method against four baselines}. Each point shows the percentage of times our approach was preferred over a given baseline, along with a 95\% Wilson confidence interval. Across all comparisons, our method is significantly preferred, indicating that it better satisfies key qualitative criteria such as realism, relevance to the prompt, seam consistency, and overall visual quality.}
  \label{fig:userstudy_results}
\end{figure}

Figure~\ref{fig:userstudy_results} summarizes the user preferences. Across all baselines, our method is consistently preferred, with a 95\% Wilson confidence interval confirming statistical significance in every case. These results indicate that our approach better satisfies the key visual quality criteria compared to existing methods.

\section{Further Limitations}

In this section, we expand on the technical and ethical limitations of our approach, complementing the discussion provided in the main paper.

As mentioned in the conclusion, our pipeline relies on two versions of the SD3 model: the first stage uses a LoRA-fine-tuned SD3 for tangent-plane generation, while the refinement stage requires the original, unmodified SD3. Consequently, both sets of weights must be available during inference.

Although our tangent-plane representation reduces the geometric distortions commonly found in equirectangular and cubemap projections, it does not eliminate them entirely, especially near the poles. In theory, increasing the number of tangent views would yield more accurate reconstructions, but doing so would introduce significant memory and computational overhead. Moreover, current generative models like SD3 remain limited in their ability to produce high-resolution content at large scale.

\begin{figure}[!t]
    \centering
    \includegraphics[width=\linewidth]{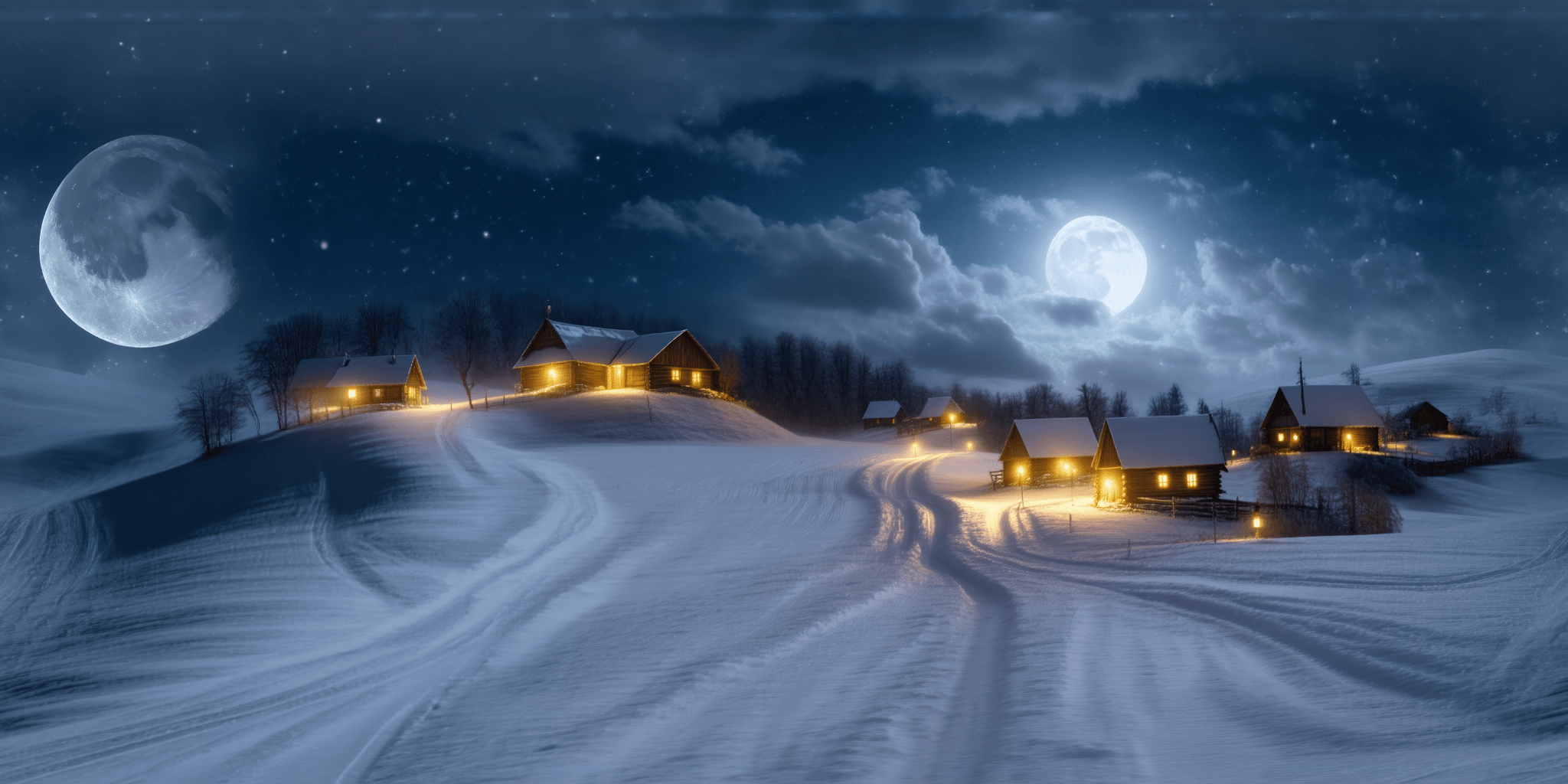}
    \caption{\textbf{A limitation of the patched denoising approach.} In this example, two patches independently generated a moon, resulting in duplicate content in the final image.}
    \label{fig:duplicate_content}
\end{figure}

To support high-resolution output (e.g., 4K panoramas), we apply a patched denoising strategy in the refinement stage, where the latent representation of the ERP image is split into smaller segments, denoised independently, and then stitched back together. While this improves scalability, it can lead to duplicate content, as each patch is conditioned on the same global caption without local awareness. As illustrated in Figure~\ref{fig:duplicate_content}, this may result in artifacts such as repeated objects, e.g. two moons generated in different patches of the sky. While the panorama remains globally consistent and structurally valid, such repetitions may deviate from the intended semantics of the prompt.

The refinement stage also remains essential for correcting inconsistencies introduced by the tangent-plane generation process. These artifacts result from the flow-matching loss used during training, which does not impose explicit geometric or pixel-wise consistency. While alternative strategies such as stronger geometric loss functions, alignment-aware objectives, or inference-time guidance could address these issues, we leave such explorations to future work.

Lastly, we acknowledge potential ethical concerns. As with other generative models, TanDiT could be misused to produce photorealistic but fictitious 360$^{\circ}$ content, potentially contributing to misinformation or deceptive media, especially in immersive applications like VR. Although this risk is not unique to our method, it highlights the importance of responsible deployment, including moderation tools, usage guidelines, and methods such as watermarking to detect synthetic content.

\section{License, Dataset, and Benchmark Release Plans}

Upon acceptance, we will publicly release the following assets to support reproducibility and foster further research in panoramic image generation:

\begin{itemize}[left=0cm]
    \item Both full and summarized captions for the Flickr360, Polyhaven, and Matterport3D datasets used in our experiments.
    \item Precomputed tangent-plane grids (in both image and latent formats) for all training and evaluation samples.
    \item A unified implementation of all evaluation metrics, including standard metrics (FID, IS, CLIP Score, OmniFID, FAED, DS) and our newly proposed {TangentFID} and {TangentIS}.
    \item Complete code for data preprocessing, training, inference, refinement, evaluation, and visualization, along with instructions and configuration files.
    \item The LoRA-fine-tuned Stable Diffusion 3.5-Large model used in our experiments, along with all necessary configuration files.
\end{itemize}

All assets will be hosted on {GitHub} (for code and documentation) and {HuggingFace} (for datasets and model weights). Captions and code will be released under a {CC-BY-NC 4.0} license. Model weights and training scripts will follow the original license of Stable Diffusion 3.5 and HuggingFace's model hub guidelines.

\end{document}